\documentclass[10pt]{article} %
\usepackage[preprint,logo]{tmlr}

\usepackage{amsmath,amsfonts,bm}

\def\eqref#1{equation~\ref{#1}}

\def\1{\bm{1}}

\DeclareMathAlphabet{\mathsfit}{\encodingdefault}{\sfdefault}{m}{sl}
\SetMathAlphabet{\mathsfit}{bold}{\encodingdefault}{\sfdefault}{bx}{n}

\usepackage{hyperref}
\usepackage{cleveref}

\usepackage{url}
\usepackage{graphicx}
\usepackage{caption}
\usepackage{booktabs}
\usepackage{subcaption}
\usepackage[percent]{overpic}
\usepackage{todonotes}

\usepackage[utf8]{inputenc}
\DeclareUnicodeCharacter{2014}{---} %
\DeclareUnicodeCharacter{2013}{--}  %
\DeclareUnicodeCharacter{2019}{'}   %
\DeclareUnicodeCharacter{201C}{``}  %
\DeclareUnicodeCharacter{201D}{''}  %

\makeatletter
\@ifundefined{HyOrg@maketitle}{}{\let\maketitle\HyOrg@maketitle}
\makeatother

\title{A Weather Foundation Model for the Power Grid}
\date{} %

\author{%
  Cristian Bodnar$^1$,
  Rapha\"{e}l Rousseau-Rizzi$^2$,
  Nikhil Shankar$^1$, James Merleau$^2$, \\ Stylianos Flampouris$^1$, Guillem Candille$^2$, Slavica Antic$^2$, \\
  Fran\c{c}ois Miralles$^{2}$,
  Jayesh~K.~Gupta$^{*,1}$ \\
  \normalfont
  \vspace{0.7em}
  \addr{$^1$Silurian AI} $\;\;\;\;\;\;\;\;$ \addr{$^2$Hydro-Qu\'ebec} \\
  \vspace{0.7em}
  \small{$^*$corresponding author: \texttt{jayesh@silurian.ai}}
}

\def\tmlr@monthtext{MM}  %
\def\tmlr@yeartext{YYYY} %
\def\openreview{\url{https://openreview.net/forum?id=XXXX}} %

\begin{document}

\makeatletter
\@maketitle\@thanks
\makeatother

\begin{abstract}
Weather foundation models (WFMs) have recently set new benchmarks in global forecast skill, yet their concrete value for the weather-sensitive infrastructure that powers modern society remains largely unexplored. In this study, we fine-tune Silurian AI's 1.5B-parameter WFM, Generative Forecasting Transformer (GFT), on a rich archive of Hydro-Québec asset observations—including transmission-line weather stations, wind-farm met-mast streams, and icing sensors—to deliver hyper-local, asset-level forecasts for five grid-critical variables: surface temperature, precipitation, hub-height wind speed, wind-turbine icing risk, and rime-ice accretion on overhead conductors. Across 6–72 h lead times, the tailored model surpasses state-of-the-art NWP benchmarks, trimming temperature mean absolute error (MAE) by $15\%$, total-precipitation MAE by $35\%$, and lowering wind speed MAE by $15\%$. Most importantly, it attains an average precision score of 0.72 for day-ahead rime-ice detection, a capability absent from existing operational systems, which affords several hours of actionable warning for potentially catastrophic outage events. These results show that WFMs, when post-trained with small amounts of high-fidelity utility data, can serve as a practical foundation for next-generation grid-resilience intelligence.
\end{abstract}

\section{Introduction}

Modern electricity networks are becoming markedly more weather sensitive. Rapid penetration of wind and solar introduces variability that is tightly coupled to mesoscale meteorology, while ageing transmission and distribution grids face mounting exposure to storms, icing, and heatwaves. Grid operators must now balance supply and demand in real time in ever more reactive way, safeguard assets against extreme events, and plan maintenance in a climate where ``once-in-a-century'' conditions recur with unsettling frequency. At the same time, the decarbonization of the economy increases social vulnerability to power outages, enhancing the need for reliability. Traditional numerical weather prediction (NWP) feeds remain indispensable, yet their spatial granularity (often $\geq$ 3 km) limits their usefulness for hyper-local, asset-level decisions such as dynamic line rating~\citep{ieee738_2013}, de-icing crew dispatch, or curtailment of turbine fleets in icing conditions.

Addressing these granular, asset-level challenges requires a new forecasting paradigm, one offered by recent advances in the world of large scale machine learning.
The foundation-model paradigm of pretraining large transformers on petabyte-scale data, followed by fine-tuning for downstream tasks, has found great success across a variety of disciplines~\citep{bommasani2021foundations,brown2020language,openai2023gpt4,radford2021clip,kirillov2023sam,jumper2021alphafold,rives2021biological,videopoet2023}.
Weather foundation models (WFMs) apply this paradigm to the task of atmospheric prediction, leveraging vast reanalysis, satellite, and other climate data archives. 
By learning flow-consistent latent representations across lead times and scales, WFMs have surpassed operational NWP in global skill metrics while offering quick inference on commodity GPUs. Crucially, their parameter sharing and attention mechanisms allow regional adaptation with orders-of-magnitude fewer labelled samples than would be required to train a model from scratch.

While the AI weather models now match or surpass traditional operational NWP models for conventional forecasting tasks at broad, continent-spanning scales~\cite{bodnar2025aurora,chantry2025aifs,graphcast2023,bi2023pangu,pathak2022fourcastnet,chen2023fuxi}, this work demonstrates a pivotal new capability brought about by WFM's regional adaptation capabilities: forecast products tailored to user needs: \emph{both} flexibly and rapidly. 
This is a difficult endeavor using existing NWP models. This makes WFM technology invaluable for utility companies since the grid isn't much impacted by ``averaged'' and generic weather variables, but rather by the ways the weather interact with assets right here, right now: on this ridge, over these wind turbines, along this particular stretch of transmission line. More precisely, grid operation requires forecasts that can meet three challenging requirements at once:
\begin{itemize}
    \item \textbf{Hyper-locality}: The forecast is sensitive enough to capture kilometer-scale pecularities like terrain, land-use, and other environmental factors in the vicinity.
    \item \textbf{Specificity}: Highly task-specific variables such as thermal ratings, renewable output, and icing risk can be forecast.
    \item \textbf{Actionability}: The forecast is timely, clear and expressed in terms that facilitate real-time decision making by grid operators (e.g., probability to exceed a critical threshold along with information on the false alarm rate).
\end{itemize}

WFMs are uniquely suited to meet these challenges, yet their potential remains largely untapped. 
This is not due to a failure of the technology, but rather a gap in awareness and a lack of established pathways for integrating their highly specific data into grid operations. This work aims to bridge this gap, showcasing how WFM technology can empower operators and build a more resilient energy infrastructure.
To illustrate WFM capabilities for the grid, in this paper we adapt Silurian's 1.5-billion-parameter WFM, Generative Forecasting Transformer (GFT) to Hydro-Québec’s asset network, producing hyper-local hourly forecasts for five grid-critical variables: surface temperature, accumulated precipitation, hub-height wind speed, wind-turbine icing risk, and rime-ice accretion on overhead conductors. Against state-of-the-art NWP baselines over 1–72 h horizons, the fine-tuned model
\begin{enumerate}
    \item Lowers temperature MAE by 15\%,
    \item Reduces total-precipitation MAE by 35\%,
    \item Reduces wind speed MAE by 15\%, and
    \item Achieves an unprecedented 0.72 average precision for day-ahead rime-ice detection, providing several additional hours of warning compared to existing process.
\end{enumerate}

These gains translate into tangible operational benefits: earlier de-icing interventions, more reliable dynamic line ratings, and improved renewable dispatch (see \Cref{sec:deicing-decision-transmission}). The study thus demonstrates that WFMs, when enriched with utility-grade observations, can become a cornerstone of next-generation grid-resilience intelligence.

\begin{figure}
    \centering
    \includegraphics[width=1.0\linewidth]{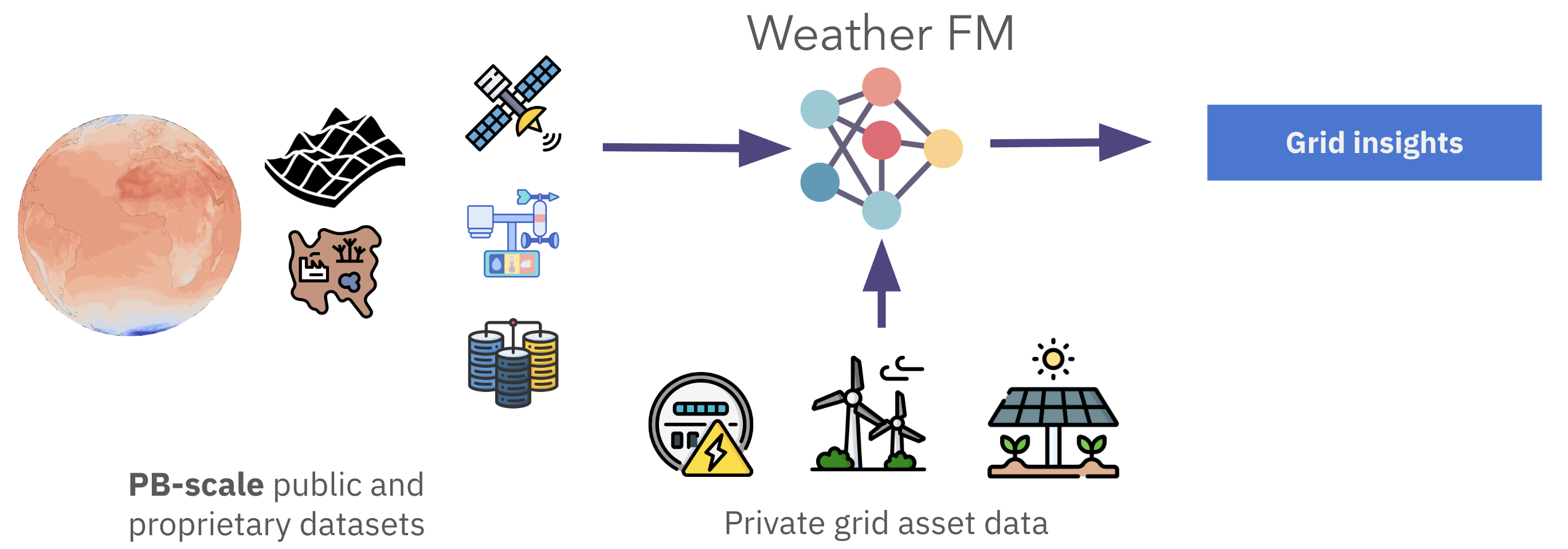}
    \caption{Weather foundation model for the power grid.}
    \label{fig:wfm_overview}
\end{figure}

\section{A Weather FM for the Grid}
\label{sec:wfm-grid}

\paragraph{Weather Foundation Models.} Recent foundation-model efforts such as ClimaX~\citep{climax23} and Aurora~\citep{bodnar2025aurora} have established that a single transformer-based backbone, pre-trained on heterogeneous climate data archives, can surpass global numerical weather prediction (NWP) skill. They also demonstrate how these models can be efficiently adapted for a wide variety of tasks in this space, ranging from wave modeling and air-pollution forecasting to regional downscaling. An overview of our WFM setup for grid applications is shown in \Cref{fig:wfm_overview}.
In this work, we show how Silurian's foundation model, the Generative Forecasting Transformer (GFT), inheriting the same principles, can be adapted for a variety of energy infrastructure use cases that demand hyperlocal spatial detail and complex weather risk assessments.

\paragraph{Architecture Overview.} The model follows the general encoder–backbone–decoder paradigm (\Cref{fig:wfm_overview}). The encoder ingests operational weather inputs such as gridded and sparse observations and maps them into a latent representation. The backbone consists of a large transformer that acts as a neural simulator of the latent representation. Finally, the decoder translates the evolved latent representations into physical predictions. The GFT decoder contains two modules: a ``dense'' decoder and a ``sparse'' hyperlocal decoder. The dense decoder produces gridded outputs on a predefined coordinate grid, while the sparse decoder produces forecasts at a specific set of query latitude–longitude locations. 

\paragraph{Pretraining and Post-training.} The model is pretrained on a petabyte-scale dataset consisting of a mix of public and private data, with dense/gridded observations forming the bulk of this dataset. The general recipe is similar to the approach given in \citet{bodnar2025aurora}. After pretraining on this large dataset, we post-train the model on a much smaller set of hyperlocal observations from sensors across the Hydro-Québec grid infrastructure. 

\paragraph{Post-training versus post-processing.} Utilities typically adapt fixed NWP guidance with Model Output Statistics, ensemble calibration, or site-specific regressors that operate one variable and one location at a time. Because the upstream flow solution never changes, these post-processing pipelines struggle to maintain multivariate consistency, cannot introduce new targets absent from the NWP feed, and scale poorly as the asset portfolio grows. Our post-training updates the core WFM weights so that a single backbone jointly forecasts temperature, wind, precipitation, and icing. The shared latent state enforces cross-variable coherence, enables the model to learn novel heads such as rime-ice probability directly from observations, and eliminates the need to maintain dozens of bespoke statistical correctors.

\paragraph{Hydro-Québec post-training setup.} At inference time, the encoder ingests ECMWF-IFS analysis fields~\citep{ecmwf_ifs_cy47r3} and the sparse Hydro-Québec asset observations collected at the cycle issue time. The decoder is trained to produce five operational targets in one forward pass: 2 m temperature, hourly precipitation, hub-height wind speed, wind-farm icing probability, and rime-ice probability on transmission lines. We fine-tune on 2016--2023 observations and hold out January 2024--March 2025 for evaluation, balancing regression losses for the continuous variables with probabilistic classification losses for the icing heads. Unless noted otherwise, utility streams provide supervision only; they are not assimilated online. 

\section{Hyperlocal forecasting for grid operations}

To demonstrate the flexibility of the foundation-model paradigm, we post-train GFT into GFT-HQ, on three important environmental tasks that have a significant impact on grid operations: (1) rime-ice forecasting for transmission lines, (2) wind-farm wind and icing forecasting, and (3) temperature and precipitation forecasting. 

\begin{figure}
    \centering
    \begin{subfigure}[b]{0.48\textwidth}
        \centering
        \includegraphics[width=\textwidth]{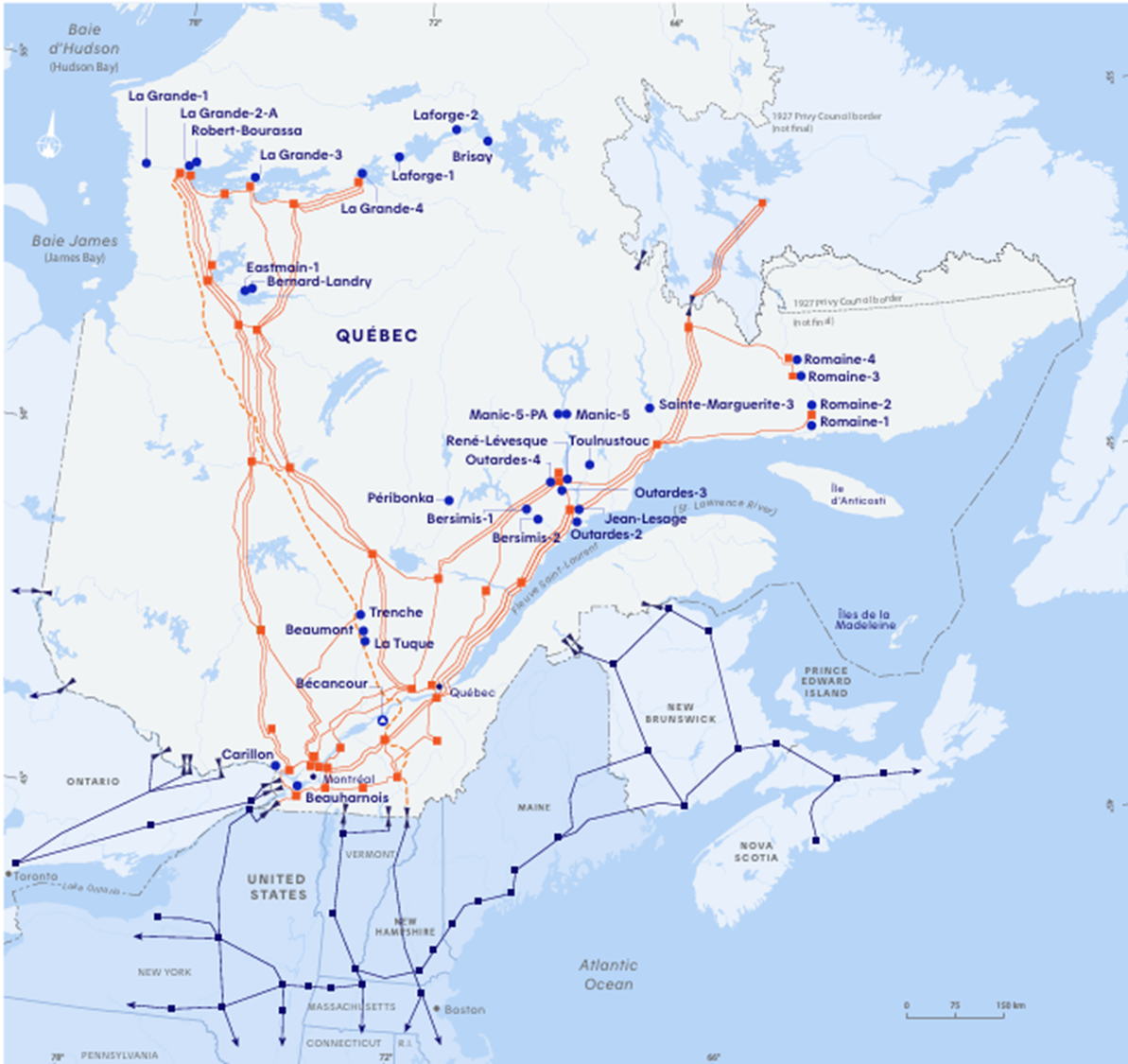}
    \end{subfigure}
    \begin{subfigure}[b]{0.48\textwidth}
        \centering
        \includegraphics[width=0.77\textwidth]{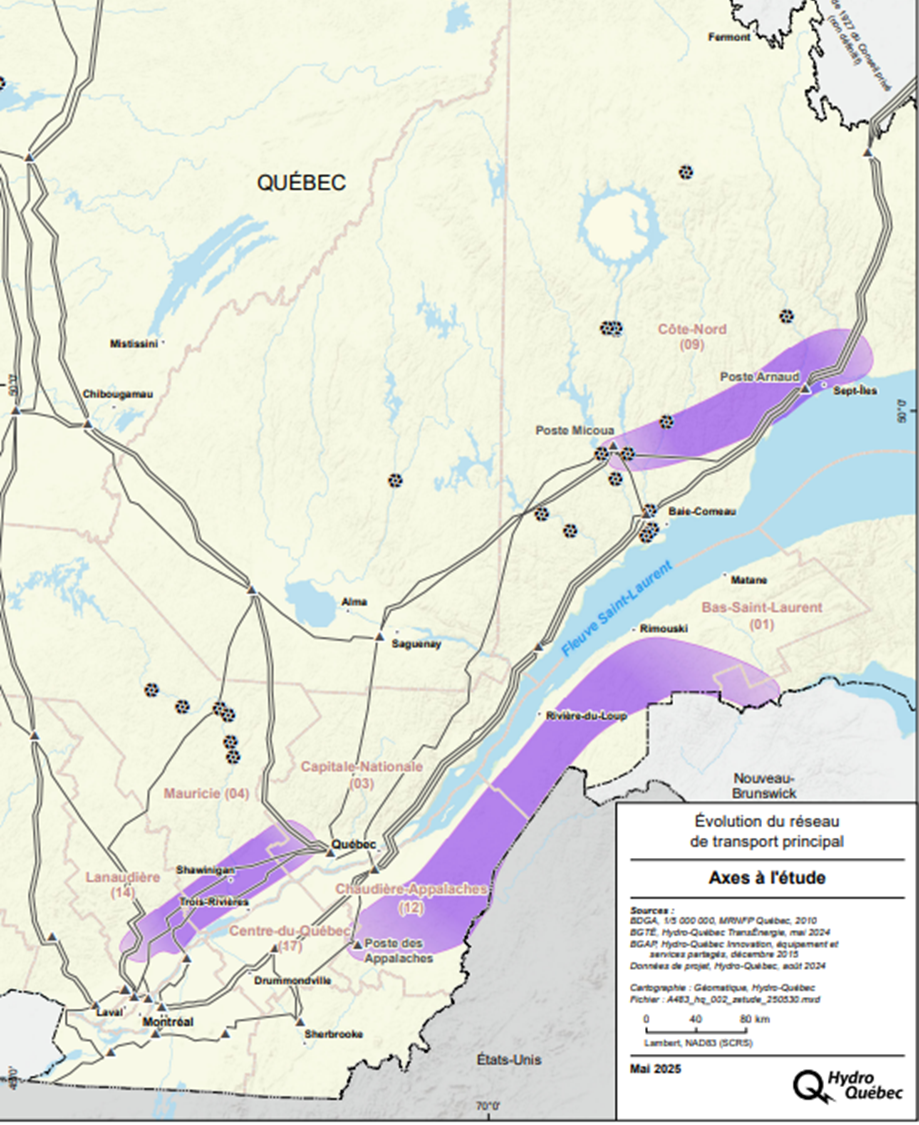}
    \end{subfigure}
    \caption{Hydro-Québec's major facilities and transmission line infrastructure: Current (Left), and planned (Right) expansion; compiled from public Hydro-Québec materials~\citep{hq_action_plan_2035_2023,hq_axes_2025,hq_supply_plan_2023,hq_strategic_plan_2024}.}
    \label{fig:hq_transmission_lines}
\end{figure}

\Cref{fig:hq_transmission_lines} illustrates Hydro-Québec's existing high-voltage network and corridors under study for future expansion, compiled from public planning materials and reports~\citep{hq_action_plan_2035_2023,hq_supply_plan_2023,hq_strategic_plan_2024,hq_axes_2025}. These large-scale buildouts and prospective wind additions motivate asset-level forecasts that are fast, location-specific, and actionable.

There are two ways hyperlocal WFM forecast products can improve grid operations. The first is to improve the skill of forecast products already in use within the utility, while retaining a similar format and role in existing operational processes. This is the case for the temperature, precipitation, and wind forecasts in the present study. The second is to meet an operational need that is not currently addressed by existing structured process in the utility, as is the case for ahead-of-time transmission-line and wind-farm rime-ice forecasts. The effort required to integrate each product--and the benefits to the utility--differ considerably between these two types. Contributions of the first type are relatively easy to integrate into the value chain, and their benefit is determined by the improvement over available forecast products. Contributions of the second type are harder to integrate because, by definition, they require the creation of new activities within the company. The potential benefit of WFMs is much greater here, as it also comes from a qualitative change to the value chain---such as being able to pre-emptively deploy an operational team to de-ice lines ahead of the start of an event. 

\paragraph{Evaluation protocol.} Continuous targets (temperature, wind, precipitation) are scored with mean absolute error (MAE) and reported as fractional skill relative to ECMWF-IFS~\citep{ecmwf_ifs_cy47r3}. Rare-event targets (icing) are assessed with precision, recall, F1, and average precision (PR-AUC), with lift computed against the relevant base rates: 3.7\% for transmission-line rime ice and 13\% for wind-farm icing. To reflect operational decision windows, we aggregate hourly probabilities into a 24 h ``any icing'' probability $q_{t}=1-\prod_{h}(1-p_{t+h})$ alongside the hour-by-hour scores. ROC-AUC is provided for completeness but we emphasize PR metrics because they better reflect the cost of false dispatches in low-base-rate regimes. 

\subsection{Rime ice forecasting for transmission lines}

Rime ice is formed when super-cooled water droplets freeze instantly upon contact with a sub-zero surface. It accumulates rapidly under fog, low-level cloud, and mountainous or coastal weather systems and poses a persistent reliability threat to overhead transmission lines. The ice accretion alters both the geometric and electrical characteristics of conductors: it increases effective diameter and surface roughness, thereby magnifying wind drag, modifying corona onset voltage, and adding eccentric mechanical loading. These translate into elevated risk of conductor clashing, flashovers, structural member fatigue, and ultimately line outages. 

\begin{figure}[t]
  \centering
  \begin{subfigure}[b]{0.48\textwidth}
    \centering
    \includegraphics[width=\textwidth]{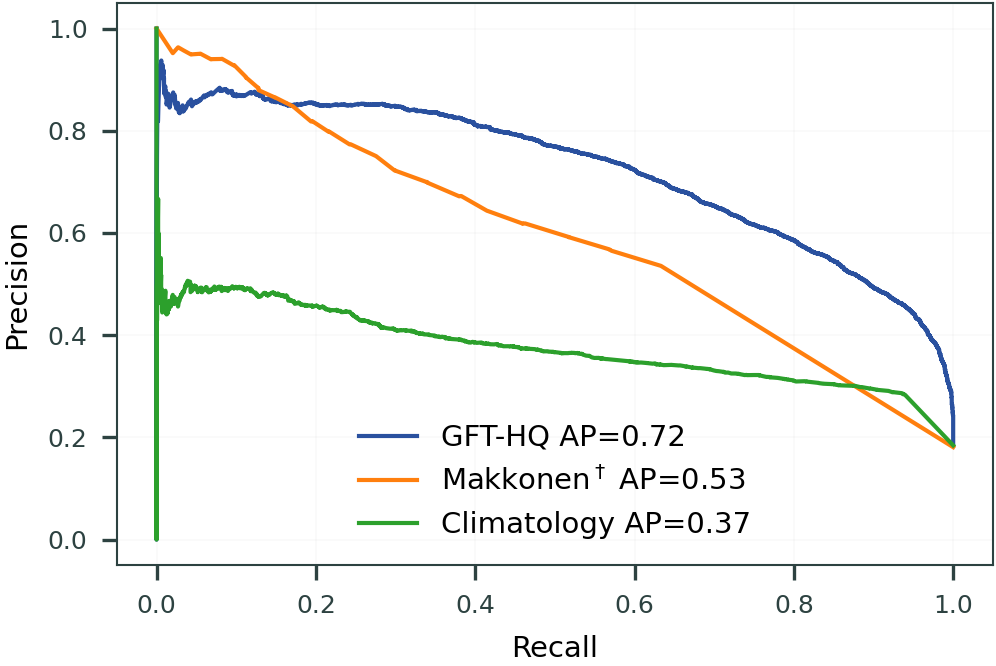}
    \label{fig:rime_ice_transmission}
  \end{subfigure}
  \begin{subfigure}[b]{0.48\textwidth}
    \centering
    \includegraphics[width=\textwidth]{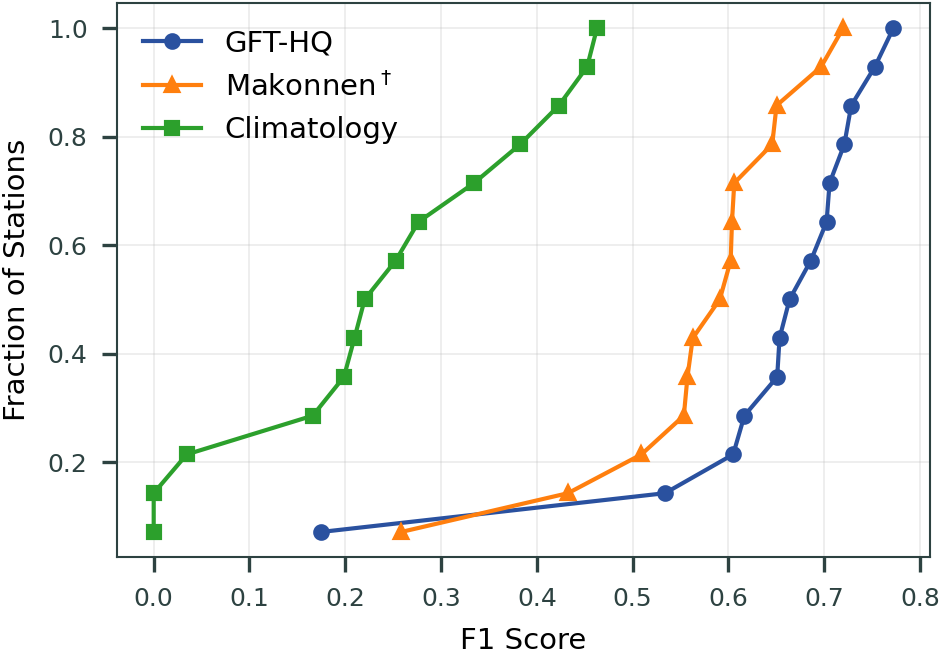}
    \label{fig:rime_ice_cdf_transmission}
  \end{subfigure}
  \caption{(a) Precision–recall curve for rime-ice forecast over the next 24 h forecasts. (b) Cumulative Distribution Function (CDF) of F1 scores of rime-ice over the next 24 h forecast over all stations. Note that Makkonen$^\dagger$ is derived from ERA5 reanalysis and cannot be used operationally.}
  \label{fig:rimeice_main}
\end{figure}

Despite the many operational headaches it causes, utilities lack reliable forecasting systems for rime ice. The gap traces chiefly to numerical weather prediction models: their kilometre-scale grids smooth out the shallow fog banks and ridge-top cloud filaments where rime forms, and their bulk microphysics schemes convert supercooled droplets to ice far too quickly, systematically erasing the liquid-water signal that drives accretion. With neither high-resolution physics nor a corridor-wide network of icing masts to assimilate or validate supercooled liquid water, model output remains both biased and unverifiable, leaving operators without a trustworthy baseline on which to build operational alerts.

\paragraph{Data}
For this task, we post-train GFT on hourly data collected from 14 high-elevation Sygivre rime-ice stations from Hydro-Québec's 40-station network; each measures ice-accretion cycles plus standard meteorological variables. See \Cref{app:datasets} for details. 

We convert the cumulative ice-accretion counts into a binary variable indicating the presence of rime ice within the last hour. In the resulting data, rime-ice events represent just $3.68\%$ of total hours, underscoring the rarity of this phenomenon and the challenge it poses for any AI-based detection system. We train on 2016-2023 and validate on January 2024 to May 2025.

\paragraph{Operational decision model} Dispatching de-icing crews carries a fixed cost $C_{d}$, while a missed icing event incurs a loss $L$ that can be partially avoided if action succeeds with effectiveness $\alpha$. Using a standard cost--loss analysis, we trigger action whenever the forecast probability exceeds
\begin{equation}
\label{eq:threshold}
p^{\star} = \frac{C_{d}}{\alpha L}.
\end{equation}
For windowed ``any icing in the next 24 h'' alerts we map hourly probabilities to $q_{t}=1-\prod_{h}(1-p_{t+h})$ and apply the same rule, with hysteresis and watch/dispatch tiers to avoid thrashing in operations. This framing connects forecast quality directly to avoided outage risk and underpins the value analysis in \Cref{sec:deicing-decision-transmission}.

\paragraph{Baselines} In addition to climatology, we compare against an ERA5-derived Makkonen index that interpolates reanalysis temperature, dewpoint, winds, and liquid-water content to each site and evaluates a physical icing proxy~\citep{hersbach2020era5}. Because this proxy is generated from retrospective reanalysis rather than a true forecast, it serves as a strong physics-based reference rather than an operational baseline.

\paragraph{Results}
Pinpointing the exact hour of icing is difficult, and hourly skill reflects this: GFT-HQ's F1 scores are typically in the 0.40–0.45 range during the first 24 h; see Appendix \Cref{fig:transmission_line_icing_hourly_f1_hourly}. Because operations often act on multi-hour windows, we also evaluate a 6-hour ``any icing in window'' target, which smooths timing errors and yields higher, more stable skill across lead times; see Appendix \Cref{fig:transmission_line_icing_hourly_f1_6h}. In both settings, GFT-HQ provides the best performance and is comparable to or better than the ERA5-derived Makkonen reference out to 72 h. As can be seen in \Cref{fig:rime_ice_transmission}, GFT-HQ also detects the approximate icing period well. For instance, when forecasting whether any icing will occur over the next 24 h at the 14 locations of interest, GFT-HQ attains average precision AP $= 0.72$ ($\approx 8\times$ lift over base rate), outperforming an ERA5-derived Makkonen index (AP $= 0.53$; $\approx 6.1\times$ lift) and climatology (AP $= 0.37$; $\approx 3\times$ lift). 
Note that Makkonen$^\dagger$ is computed from ERA5 reanalysis fields (see \Cref{app:makkonen}; ERA5 described in \citealp{hersbach2020era5}), so it is closer to observations than to a true forecast; we include it as a strong physics-based reference rather than an operational baseline. GFT-HQ exceeds this reference by $\approx 30\%$ relative AP (0.72 vs 0.53). Across 14 sites, station-level F1 scores are consistently higher for GFT-HQ, indicating improvements at the median and in the worst-case stations, not just at a few outliers.

\begin{figure}
    \centering
    \begin{subfigure}[b]{\textwidth}
    \centering
    \begin{overpic}[width=0.9\linewidth]%
      {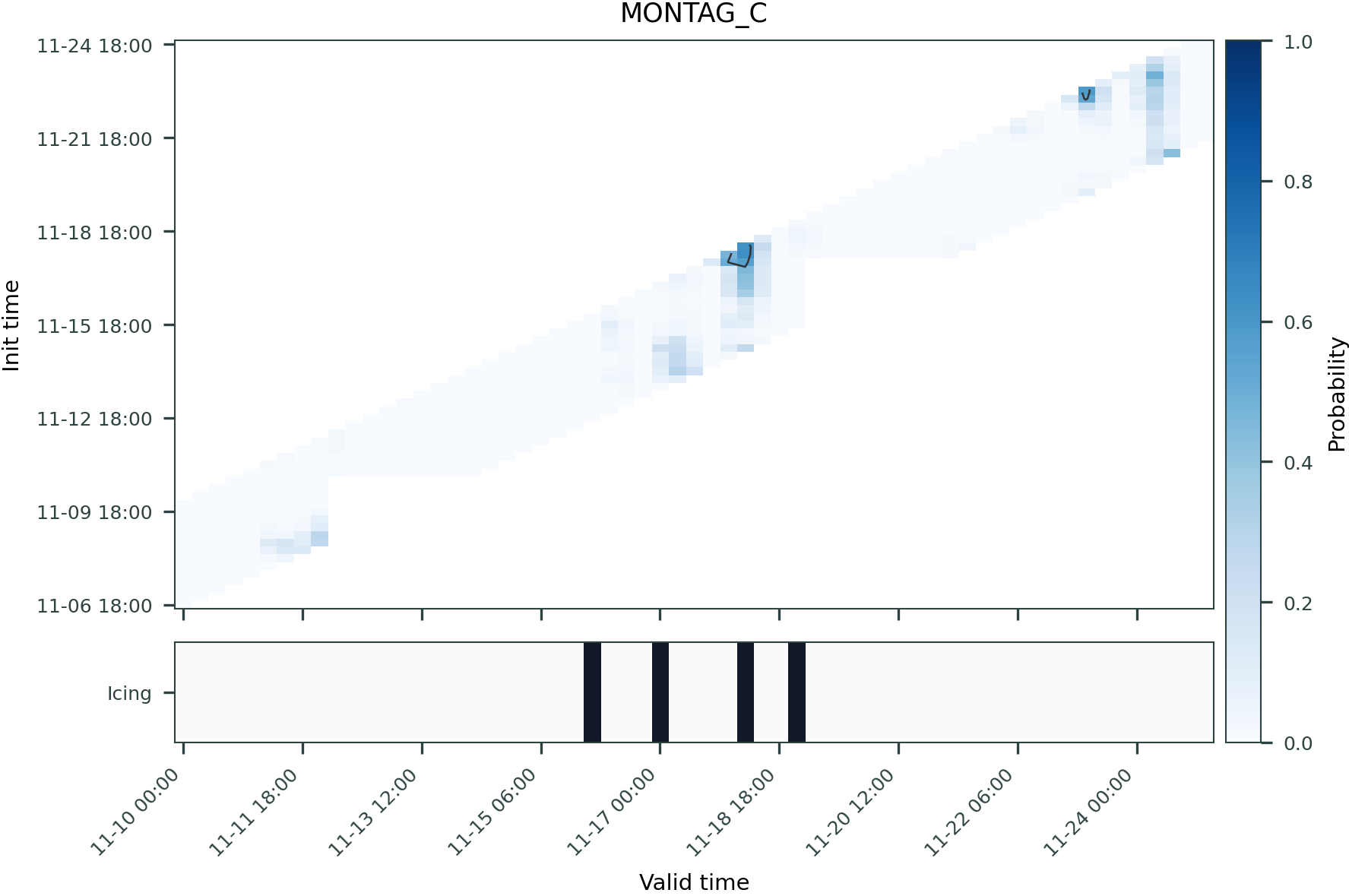}
    \end{overpic}
    \end{subfigure}
    \begin{subfigure}[b]{\textwidth}
    \centering
    \includegraphics[width=0.9\linewidth]%
    {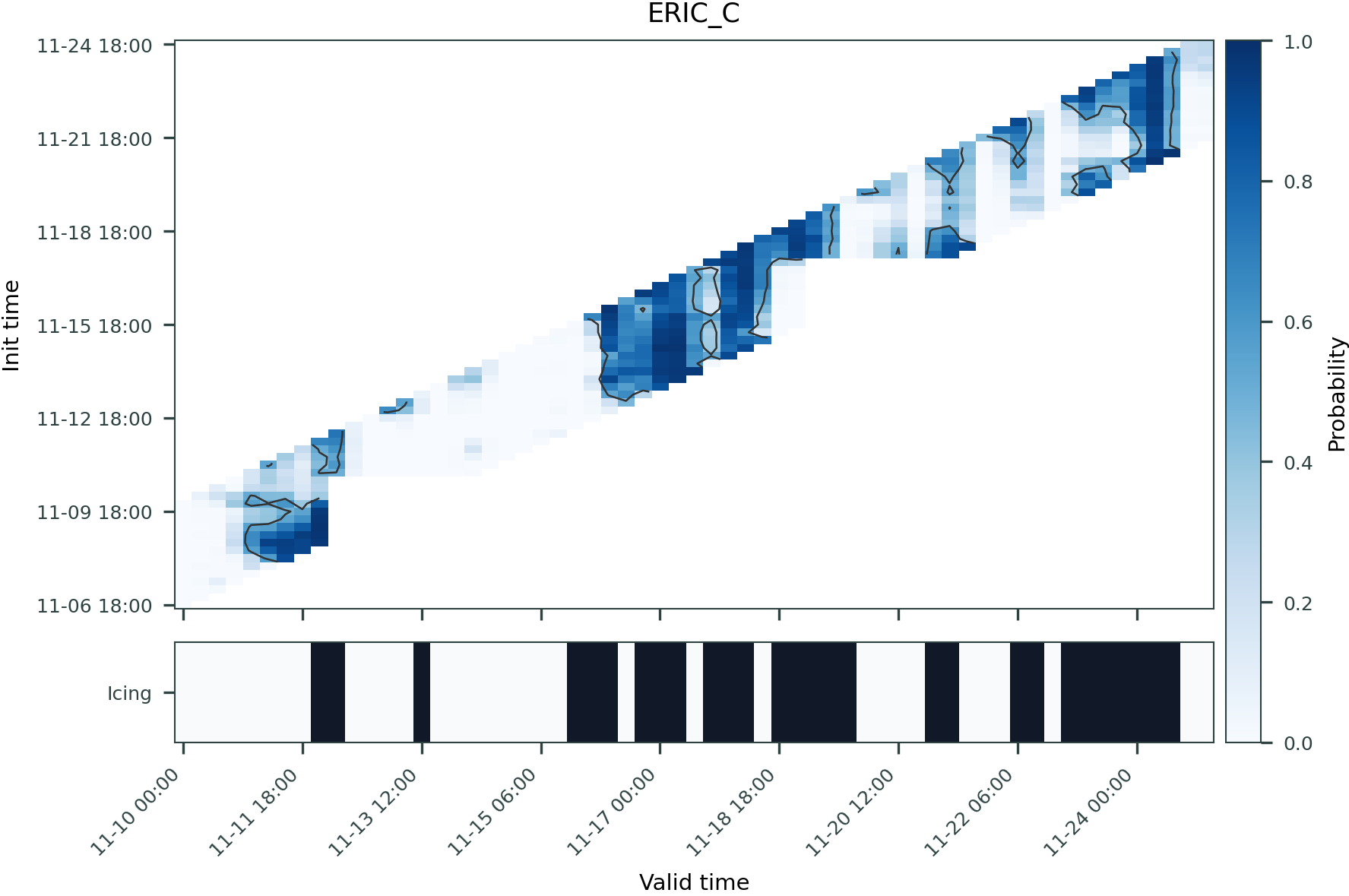}
    \end{subfigure}
    \caption{Romaine rime-ice event (Nov 2024) at two nearby Sygivre stations ~11 km apart. Top: \texttt{MONTAG\_C}; bottom: \texttt{ERIC\_C}. Heatmap shows GFT-HQ 1-h rime-ice probabilities from successive 6-hourly initializations (\emph{y}-axis) verifying at each valid time (\emph{x}-axis; UTC) over 2024-11-06 to 2024-11-24. The black strip labelled ``Icing'' is the observed binary occurrence. At \texttt{ERIC\_C}, a stable high-probability signal appears by 16~Nov and peaks on 18--19~Nov, aligning with the longest observed episode; several later bursts are also forecast. At \texttt{MONTAG\_C}, the signal is weaker and more episodic, with only modest probabilities near the brief observed bursts, illustrating strong micro-scale differences despite close proximity.}
    \label{fig:romaine_icing_event_nov_2024}
\end{figure}

\paragraph{Case study}
Beyond aggregate skill, the finetuned model captured individual high-impact episodes that posed particular challenges for Hydro-Québec's operations. 
A prime example is the collapse of a transmission line from the Romaine hydro centre around November 19, 2024 due to rime ice \citep{Gerbet2024}. 
\Cref{fig:romaine_icing_event_nov_2024} compares two nearby Sygivre stations (\texttt{ERIC\_C} and \texttt{MONTAG\_C}) separated by only ~11 km using a Hovmöller-style view: rows are successive 6-hourly forecast initializations, columns are valid hours, colours show the forecast probability of 1-h rime ice, and the black strip below indicates the observed binary icing. At \texttt{ERIC\_C}, a coherent high-probability swath is already present by 16~Nov and persists across cycles, peaking on 18--19~Nov when the longest observed episode occurs; subsequent bursts on 20--24~Nov are also captured. At \texttt{MONTAG\_C}, icing was shorter and more intermittent, and the model response is correspondingly weaker—probabilities remain low most of the time with brief increases around 16--19~Nov. Despite their proximity, these stations exhibit markedly different behaviour, underscoring sharp micro-scale variability along the corridor; nevertheless the corridor-scale risk around 18--19~Nov was visible 1--3 days in advance, providing actionable lead time for grid operations. %

\paragraph{Initialization and inputs}
Unless otherwise stated, the case-study forecasts are produced by 6-hourly GFT-HQ cycles (issue times 00, 06, 12, 18 UTC). For each cycle, the encoder is conditioned on dense ECMWF-IFS Analysis fields~\citep{ecmwf_ifs_cy47r3} (0 h analyses) at the issue time—including multi-level temperature, humidity, and winds; surface pressure and near-surface fields; cloud liquid-water proxies; static orography and land–sea mask; and time encodings. 
These fields provide the initial conditions from which the backbone evolves the latent state forward in time. Hydro-Québec asset streams (Sygivre icing stations and wind-farm masts) are used for post-training supervision; unless explicitly noted, they are not assimilated at runtime during inference. The traces in \Cref{fig:romaine_icing_event_nov_2024} therefore reflect differences solely from the changing synoptic initial state between successive 6-hourly initializations.

\begin{figure}[t]
\centering
\begin{subfigure}[b]{0.32\textwidth}
\centering
\includegraphics[width=\textwidth]{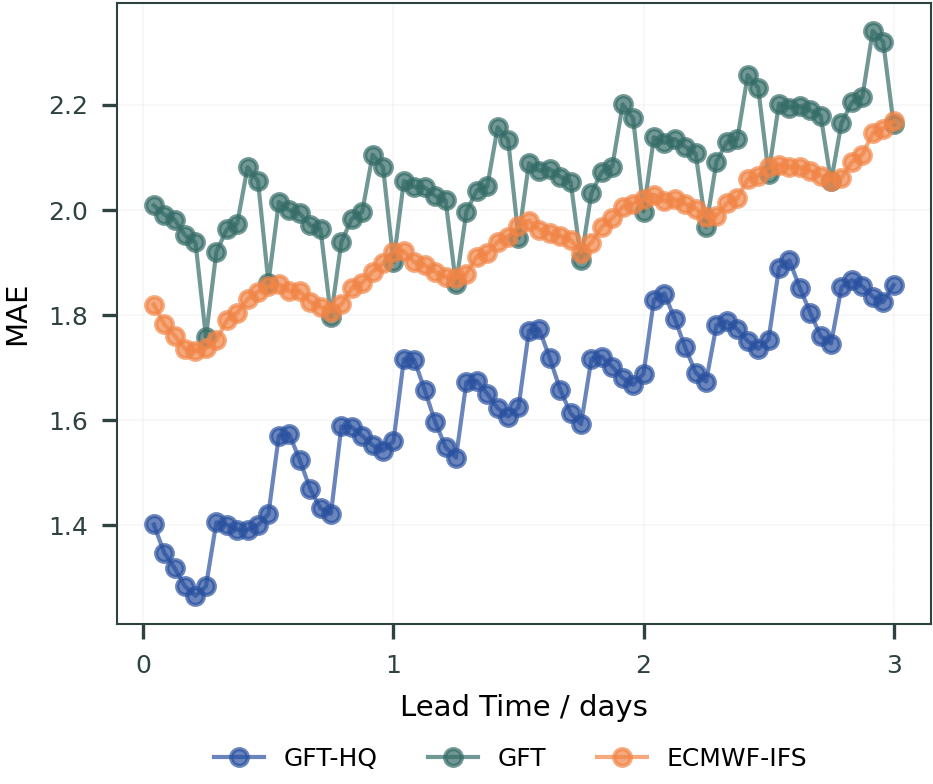}
\label{fig:windfarm_mae}
\end{subfigure}
\begin{subfigure}[b]{0.32\textwidth}
\centering
\includegraphics[width=\textwidth]{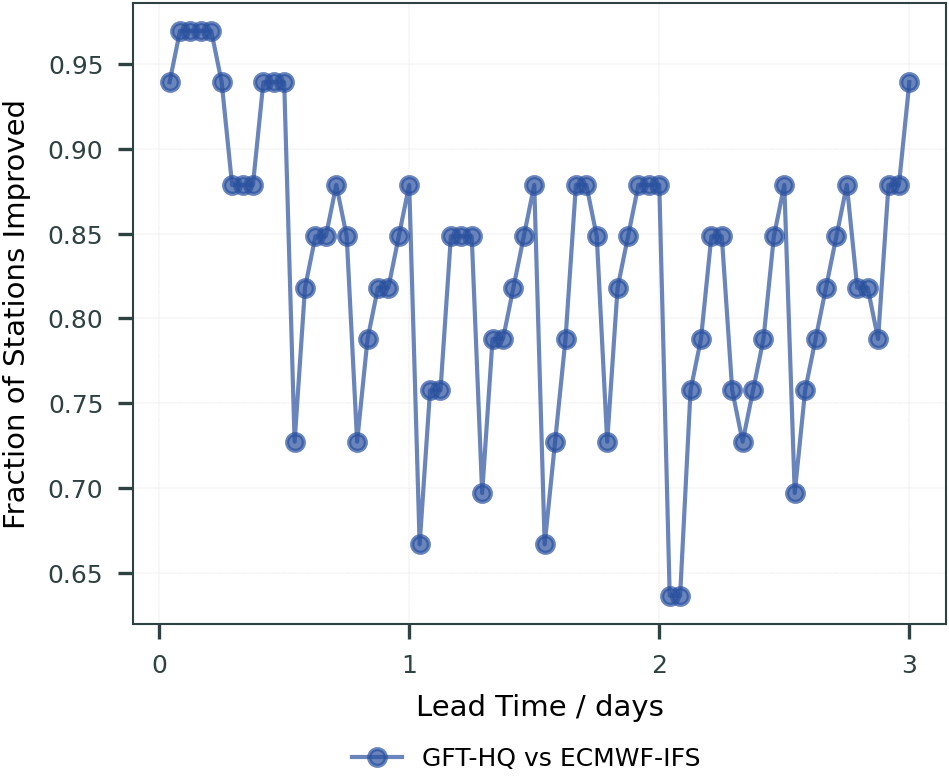} 
\label{fig:windfarm_fraction}
\end{subfigure}
\begin{subfigure}[b]{0.32\textwidth}
\centering
\includegraphics[width=\textwidth]{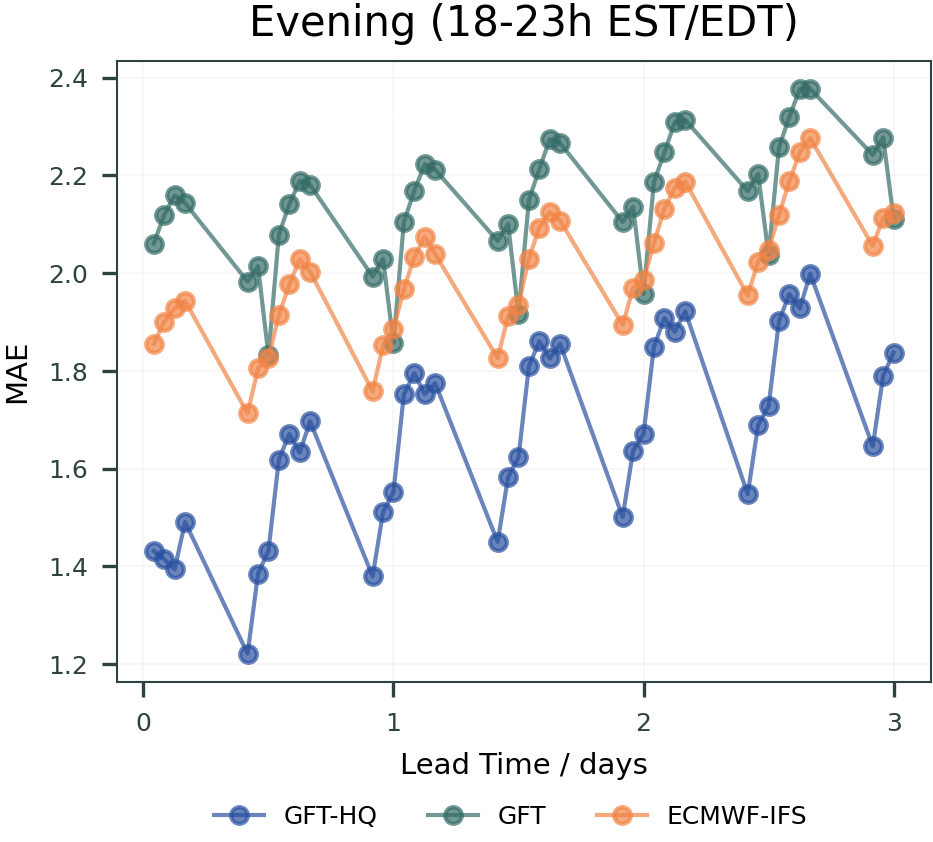} 
\label{fig:windfarm_evening_mae}
\end{subfigure}
\caption{Wind-speed forecast improvements from finetuning. (a) Mean absolute errors of forecasts across all stations (b) Fraction of stations with lower forecasting errors than ECMWF-IFS (c) Mean absolute error across all the stations in the evening during peak load }
\end{figure}

\begin{figure}
    \centering
    \begin{subfigure}[b]{0.48\textwidth}
    \phantomsubcaption\label{fig:windfarm_icing_mae} %
    \begin{overpic}[width=\textwidth]{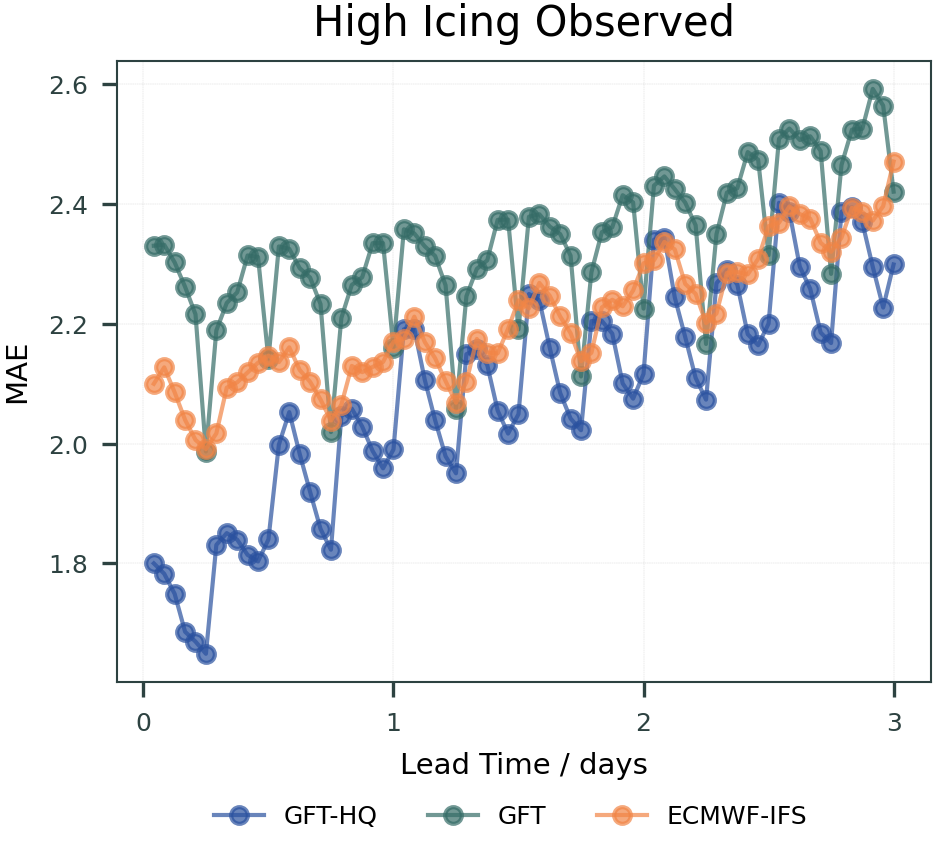}
      \put(2,82){\footnotesize\bfseries(\thesubfigure)}
    \end{overpic}
    \end{subfigure}    
    \begin{subfigure}[b]{0.48\textwidth}
    \phantomsubcaption\label{fig:windfarm_icing_csi} %
    \begin{overpic}[width=\textwidth]{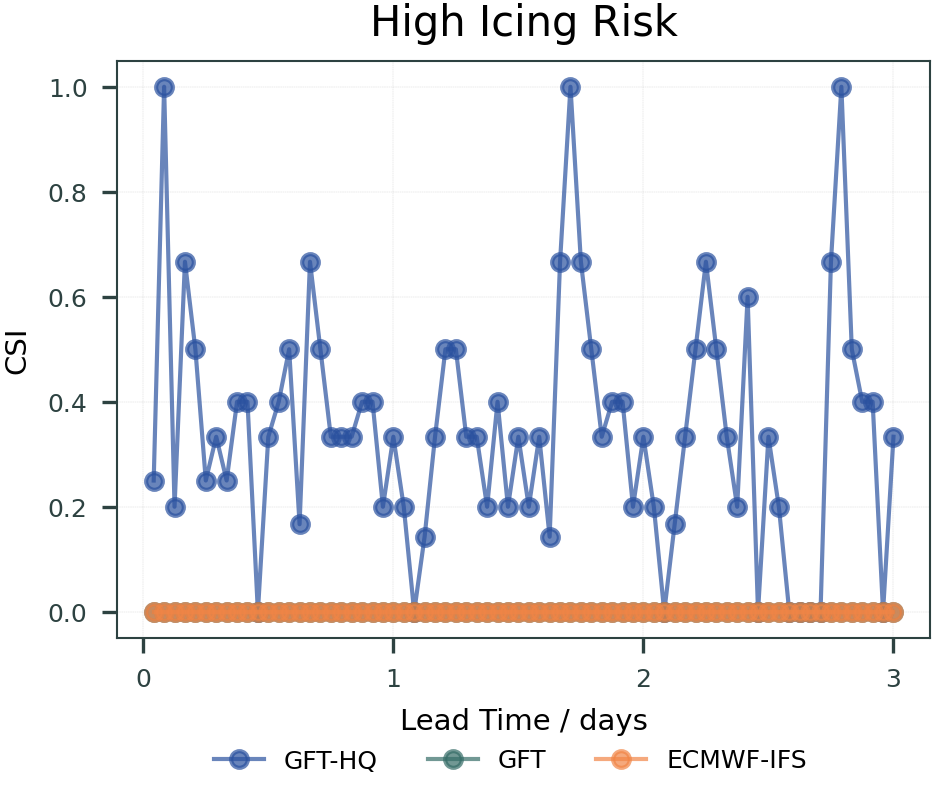}
      \put(2,82){\footnotesize\bfseries(\thesubfigure)}
    \end{overpic}
    \end{subfigure}        
    \caption{(a) Mean absolute error (m/s) across all stations during peak icing as measured by reduction in power production (b) Cut out windspeed ($> 25$m/s) detection during high icing}
    \label{fig:windfram_ws_icing}
\end{figure}

\begin{figure}[t]
  \centering
  \begin{subfigure}[b]{0.48\textwidth}
    \centering
    \includegraphics[width=\textwidth]{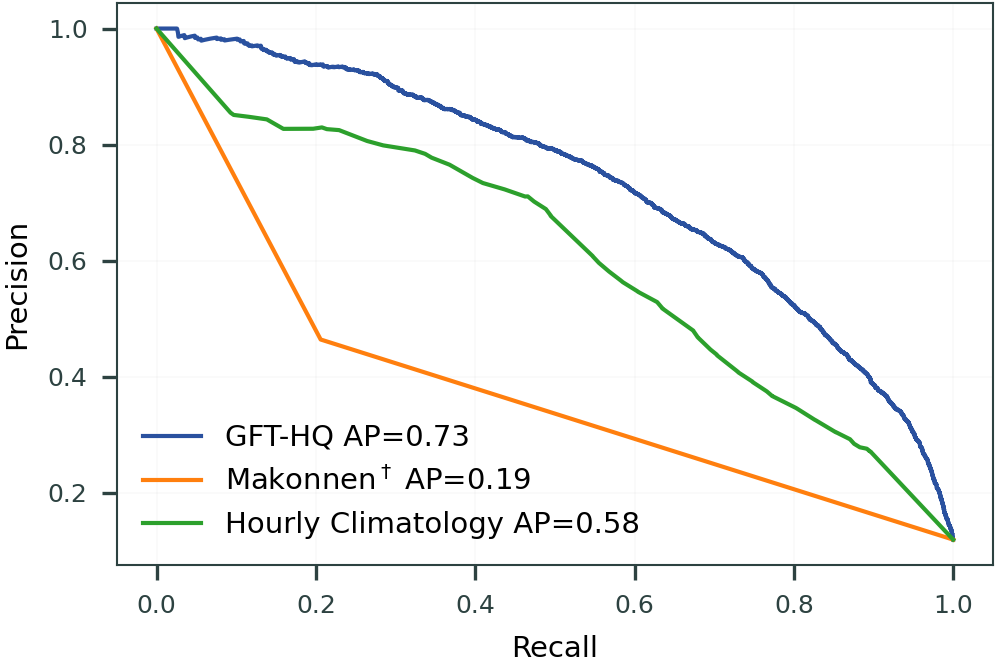}
    \caption{Forecast at 24h lead-time}
    \label{fig:sub2}
  \end{subfigure}
  \begin{subfigure}[b]{0.48\textwidth}
    \centering
    \includegraphics[width=\textwidth]{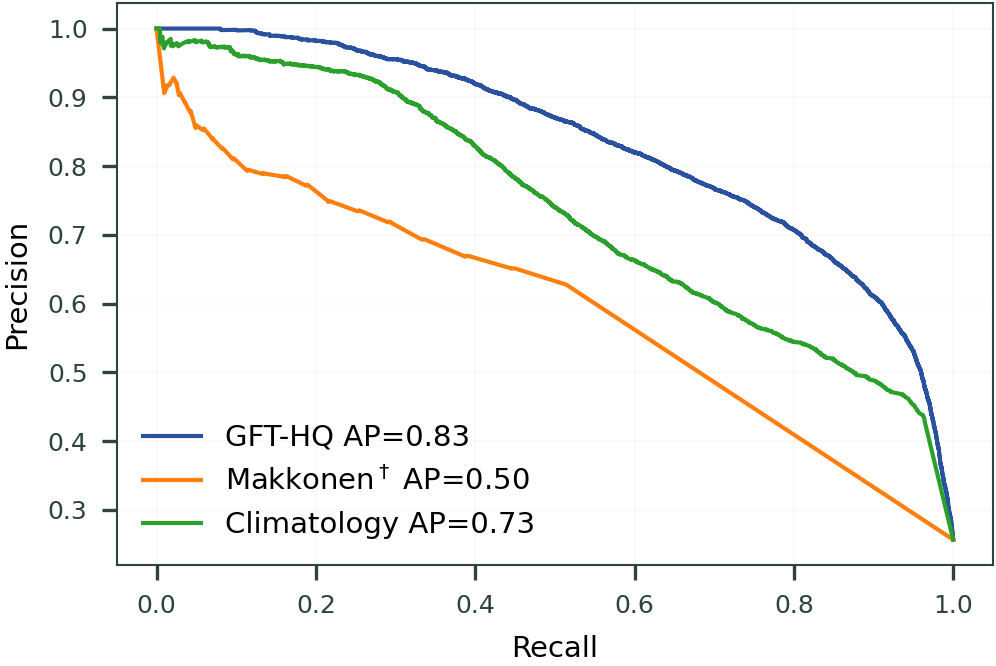}
    \caption{Forecast over 24h time-horizon}
    \label{fig:sub3}
  \end{subfigure}
  \caption{Precision–recall curves for detecting wind-turbine icing across the wind-farm dataset. The fine-tuned model (GFT-HQ) is compared against climatology and an ERA5-derived Makkonen index. Panel (a) evaluates a fixed 24 h lead; panel (b) evaluates the windowed target of any icing within the next 24 h. Average precision (AP) summarizes performance; the evaluation base rate is 13\%, so improvements correspond to substantial lift over random. Note that Makkonen$^{\dagger}$ is included as a physics-based reference rather than an operational baseline.}
  \label{fig:windfarm_ice_pr}
\end{figure}

\subsection{Wind-farm wind speed and ice forecasting}

For utilities reliable forecasts of turbine-level icing and hub-height wind speed are mission-critical because they dictate how much variable generation will actually reach the grid, how much spinning reserve must be scheduled, and whether transmission lines risk overloads or voltage excursions when production suddenly collapses. An unexpected icing episode can slash a wind-plant’s output by tens of megawatts within minutes, while excessive winds force cut-out shutdowns; both events create sharp ramps that the utility must offset with fast-responding thermal units, battery assets, or market purchases, often at premium prices. Day-ahead and intraday visibility into these weather-driven outages therefore underpins accurate load-generation balancing, congestion management, and reliable commitment of ancillary services, ultimately lowering imbalance penalties and safeguarding system stability.

\paragraph{Data}
As the grid operator, Hydro-Québec does not have direct measurements of icing on wind-turbine blades. Instead, this label is inferred from substantial power-production declines observed at a wind farm given the measured wind speed and direction. The wind speed measurements come from met masts placed next to the wind farms such that potential wake effects are negligible. As before, we fine-tune the model on data from 2016–2023 and test on data from January 2024 to January 2025. 

We use 65 meteorological masts near wind farms across Québec (10-minute data, multiple heights) and infer hourly wind-farm icing from production losses relative to potential output given the meteorological conditions. See \Cref{app:datasets} for details.

\begin{figure}[t]
  \centering
  \begin{subfigure}[b]{0.48\textwidth}
   \phantomsubcaption\label{fig:wind_diurnal_evening_mae} %
    \begin{overpic}[width=\textwidth]{figs/windfarm/windspeed_Evening_mae.png}
      \put(2,85){\footnotesize\bfseries(\thesubfigure)}
    \end{overpic}  
  \end{subfigure}
  \begin{subfigure}[b]{0.48\textwidth}
    \centering
  \phantomsubcaption\label{fig:windfarm_night_mae} %
    \begin{overpic}[width=\textwidth]{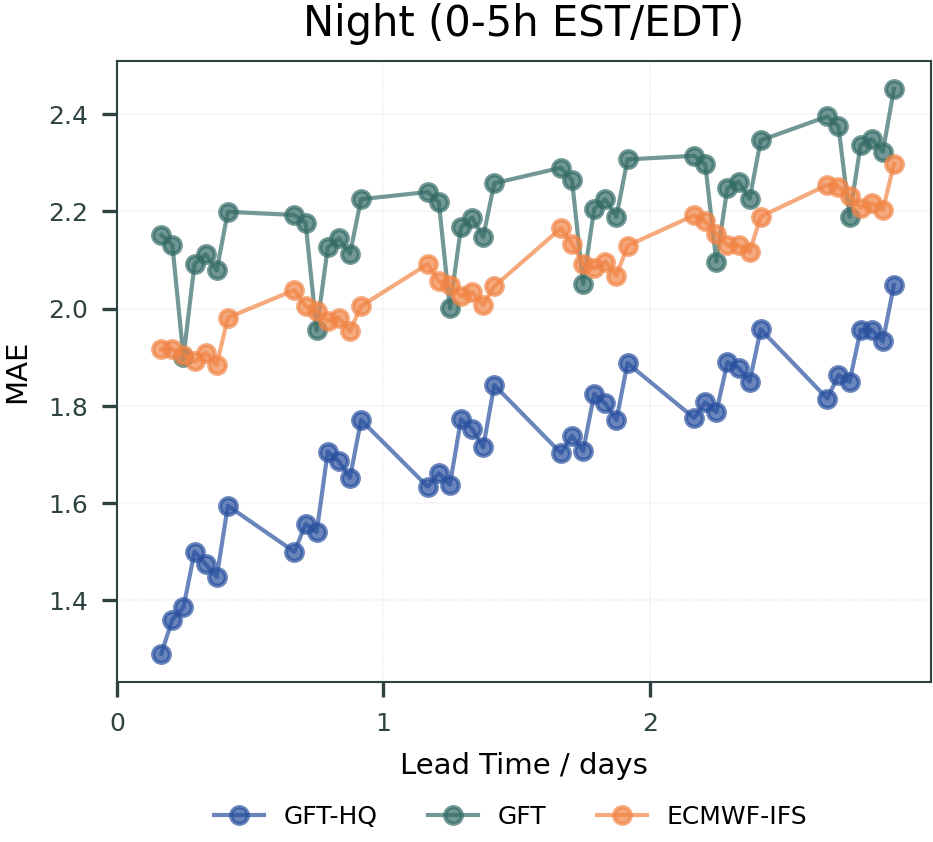}
      \put(2,85){\footnotesize\bfseries(\thesubfigure)}
    \end{overpic}      
  \end{subfigure}
  \caption{Diurnal hub-height wind MAE (m/s). GFT-HQ significantly narrows the mean absolute error during the evening/night periods when Hydro-Québec faces the highest load and icing exposure.}
  \label{fig:wind_diurnal_main}
\end{figure}
\begin{figure}[t]
    \centering
    \includegraphics[width=0.95\linewidth]{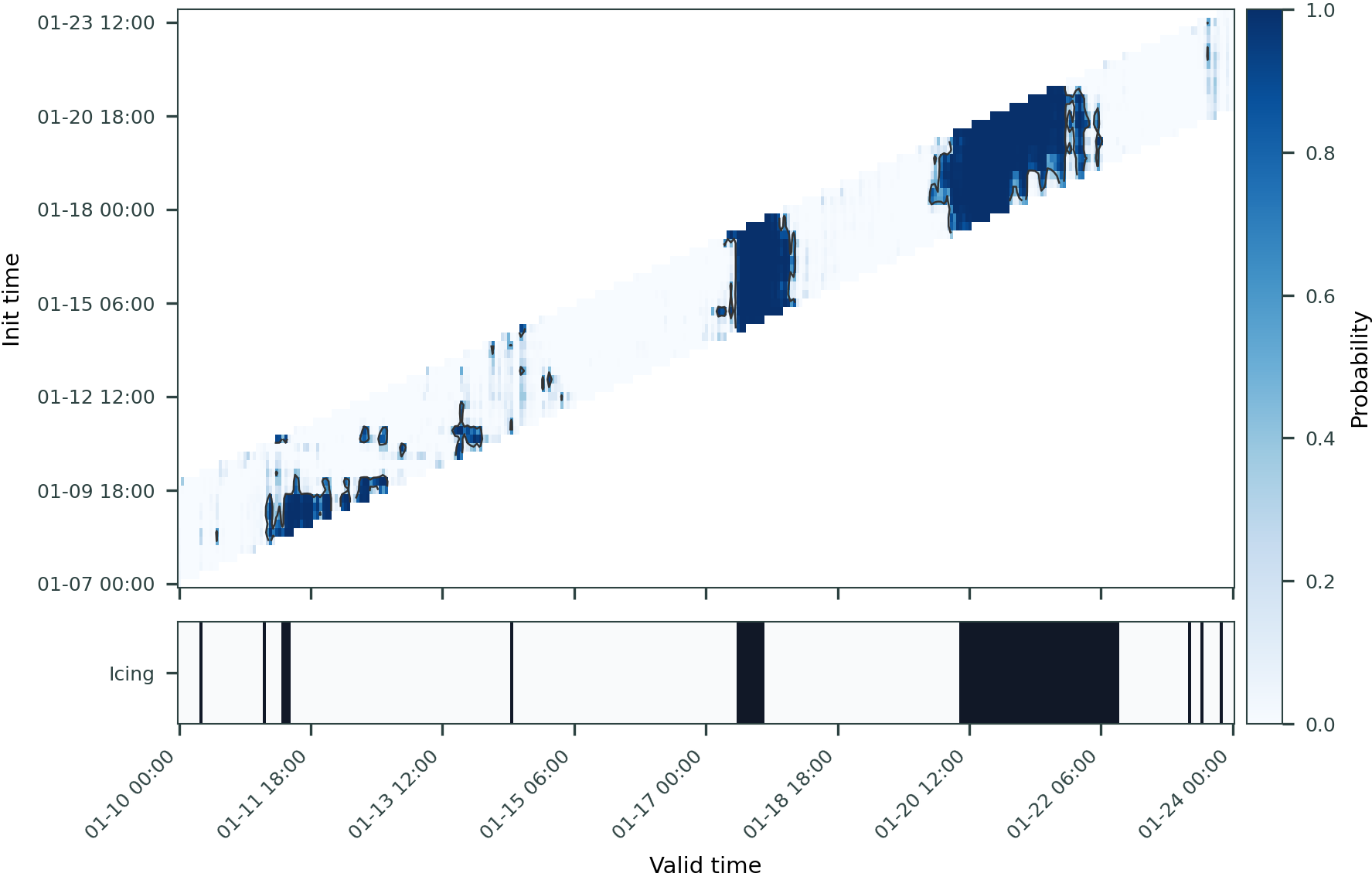}
    \caption{Wind-farm icing case study at an anonymized site (``AVA''). Rows are 6-hourly forecast initializations (\emph{y}-axis); columns are valid hours (UTC, \emph{x}-axis). The heatmap shows GFT-HQ probabilities of 1-h turbine icing; the black bar labelled ``Icing'' indicates observed occurrence inferred from production. A stable high-risk corridor appears several days in advance and peaks during the multi-day episode in late January, aligning with observations while also flagging shorter earlier bursts.}
    \label{fig:windfarm_icing_ava_casestudy}
\end{figure}

\paragraph{Results}
Across the held-out period , the finetuned model (GFT-HQ) delivers strong and geographically consistent skill on both the \emph{drivers} (hub-height wind) and the \emph{impact} target (wind-farm icing).

\textbf{Icing risk}
At a base rate of 13\%, GFT-HQ attains AP $= 0.73$ ($\approx \times$lift over random), for detecting whether \emph{any} icing will occur in the next 24 h, exceeding both Climatology (AP $=0.62$) and an ERA5-derived Makkonen index (AP $=0.50$). Evaluated strictly at 24 h lead time, GFT-HQ again leads (AP = 0.73) versus Climatology (AP=$0.62$), and Makkonen (AP $= 0.19$). Corresponding ROC performance is high as well (AUC = 0.93 vs 0.86–0.87 (Climatology) and 0.59–0.71 (Makkonen) in the two setups).
\emph{Note.} The Makkonen curve is computed from ERA5 reanalysis (see \Cref{app:makkonen}) and is included as a strong physics-based \emph{reference}, not an operational forecast baseline. Performance here is often worse than Climatology partially because icing is not directly measured at wind-farms, rather inferred from loss in power production and could include other factors as well.

\textbf{Wind-speed skill}
Finetuning materially reduces hub-height wind MAE across 0–72 h relative to untuned GFT and ECMWF-IFS, with the majority of sites showing lower error than IFS at each lead time. Gains persist and are most pronounced during evening hours (18–23 EST/EDT), which are operationally critical for peak load.

Lead-time averages remain high. Icing F1 stays above 0.6 through 60 h and ROC-AUC remains above 0.9 (plots in \Cref{app:windfarm-forecasting})—while the diurnal slices in \Cref{fig:wind_diurnal_main} illustrate how GFT-HQ damps the errors compared to both baselines. Seasonal panels are provided in \Cref{app:windfarm-forecasting}; the evening/night margins are the largest, aligning with Hydro-Québec's highest-risk periods.

\textbf{Performance under high-icing conditions.}
As can be seen in \Cref{fig:windfram_ws_icing}, conditioned on hours flagged as high icing risk, GFT-HQ keeps the wind-speed accuracy edge and is the only model with consistently non-zero skill for turbine cut-out events ($> 25$m/s). Its CSI shows frequent high-skill episodes, whereas both GFT and ECMWF-IFS are near zero.

\paragraph{Case study}
To illustrate how the system presents actionable guidance to operators, \Cref{fig:windfarm_icing_ava_casestudy} shows a mid-winter icing episode at a representative Hydro-Qu\'ebec wind farm (site code ``AVA''). The Hovm\"oller-style plot stacks successive 6-hourly forecasts by initialization time (\emph{y}-axis) against their valid hours (\emph{x}-axis); colours denote the GFT-HQ probability of 1-h turbine icing, while the black strip beneath marks the observed binary icing derived from production losses. A coherent high-probability swath emerges and stabilizes ~2--3 days before the prolonged mid-to-late January event, and shorter early-month bursts are also indicated. The cross-cycle persistence of this signal provides dependable early warning for curtailment planning and reserve scheduling.

\subsection{Temperature and precipitation forecasting}
Accurate temperature and precipitation forecasts sit at the heart of utility planning because they govern every major balance-sheet variable: demand, supply, and asset risk. Temperature is the dominant driver of electric load—hot spells boost air-conditioning demand, cold snaps spike heating and resistive losses—so even a 1 °C error can translate into hundreds of megawatts of forecasting miss and costly imbalance charges. Precipitation matters just as much: rain, snow, or ice determine hydro-reservoir inflows, solar panel soiling, wind-turbine icing shutdowns, and outage probabilities from lightning or tree-fall. By anticipating these weather-dependent swings, a utility can commit generation and reserves economically, schedule maintenance when both load and storm risk are low, pre-stage crews and mobile transformers ahead of severe weather, and optimise fuel and emissions strategies. In short, precise temperature and precipitation outlooks convert meteorological uncertainty into actionable lead time, protecting reliability while trimming operating costs.

\begin{figure}[t]
  \centering
  \begin{subfigure}[b]{0.9\textwidth}
    \centering
    \includegraphics[width=\textwidth]{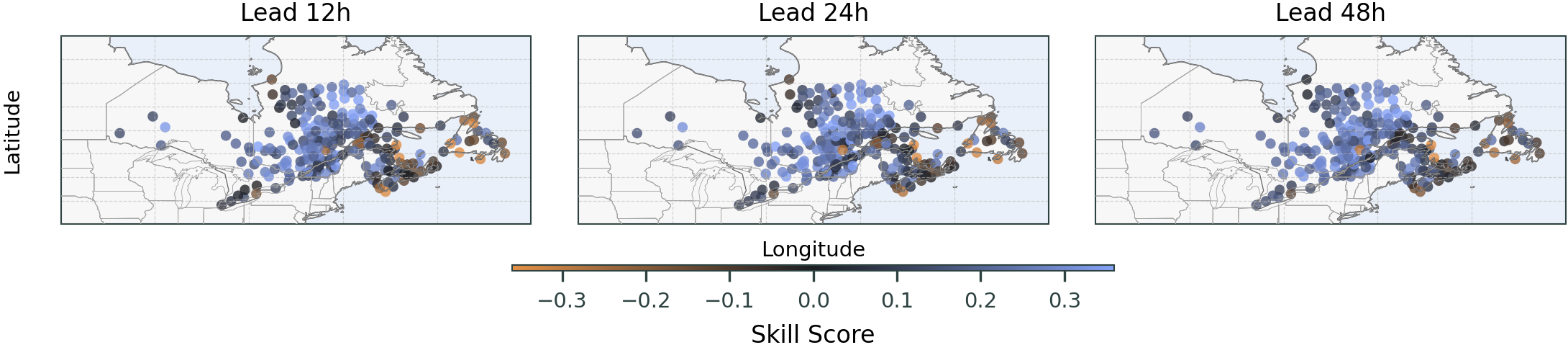}
    \caption{GFT-HQ Temperature skill vs. ECMWF-IFS}
  \end{subfigure}
  \begin{subfigure}[b]{0.9\textwidth}
    \centering
    \includegraphics[width=\textwidth]{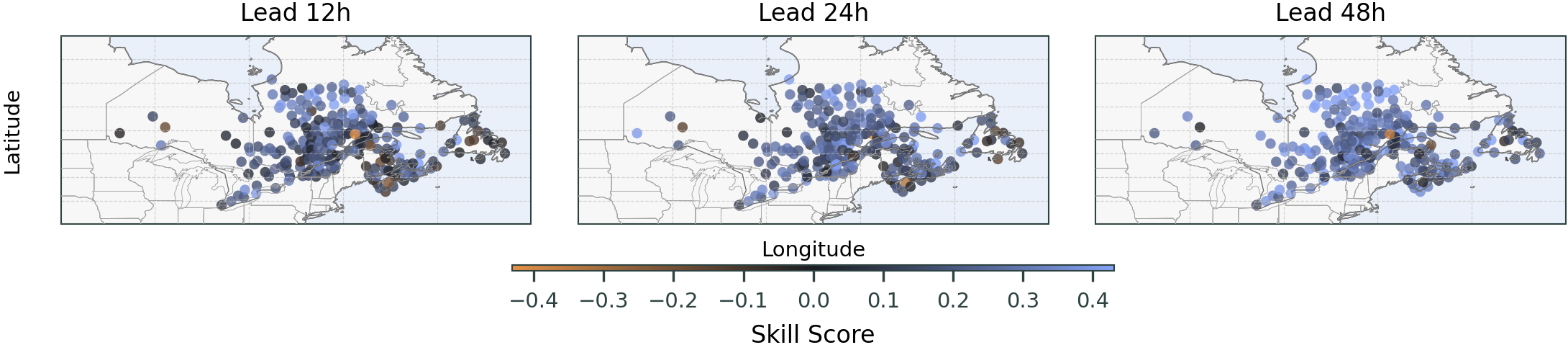}
    \caption{GFT-HQ Precipitation skill vs. ECMWF-IFS}
  \end{subfigure}
  \caption{Spatial coverage of positive fractional skill for GFT-HQ relative to ECMWF-IFS at 12 h, 24 h, and 48 h. Finetuning expands the footprint of improved stations for both temperature and hourly precipitation, supporting system-wide load and hydro inflow decisions.}
  \label{fig:temp_precip_maps_main}
\end{figure}

\begin{figure}[t]
  \centering

  \begin{subfigure}[b]{0.48\textwidth}
    \phantomsubcaption\label{fig:snap_temperature_mae} %
    \begin{overpic}[width=\textwidth]{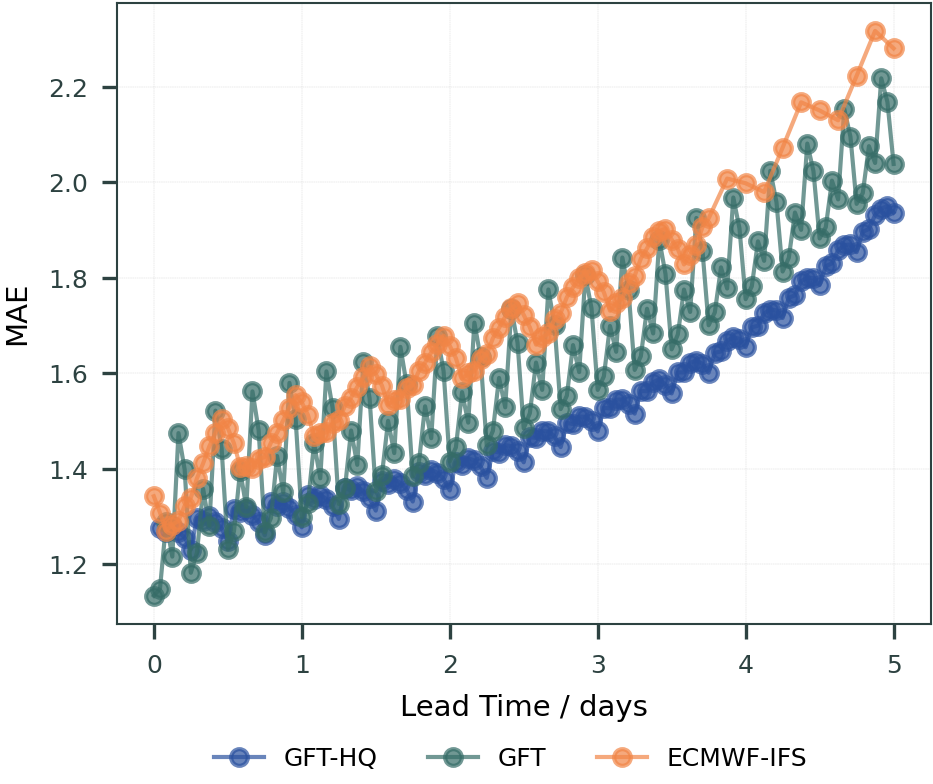}
      \put(2,82){\footnotesize\bfseries(\thesubfigure)}
    \end{overpic}
  \end{subfigure}
  \begin{subfigure}[b]{0.48\textwidth}
    \phantomsubcaption\label{fig:snap_temperature_stations_improved}
    \begin{overpic}[width=\textwidth]{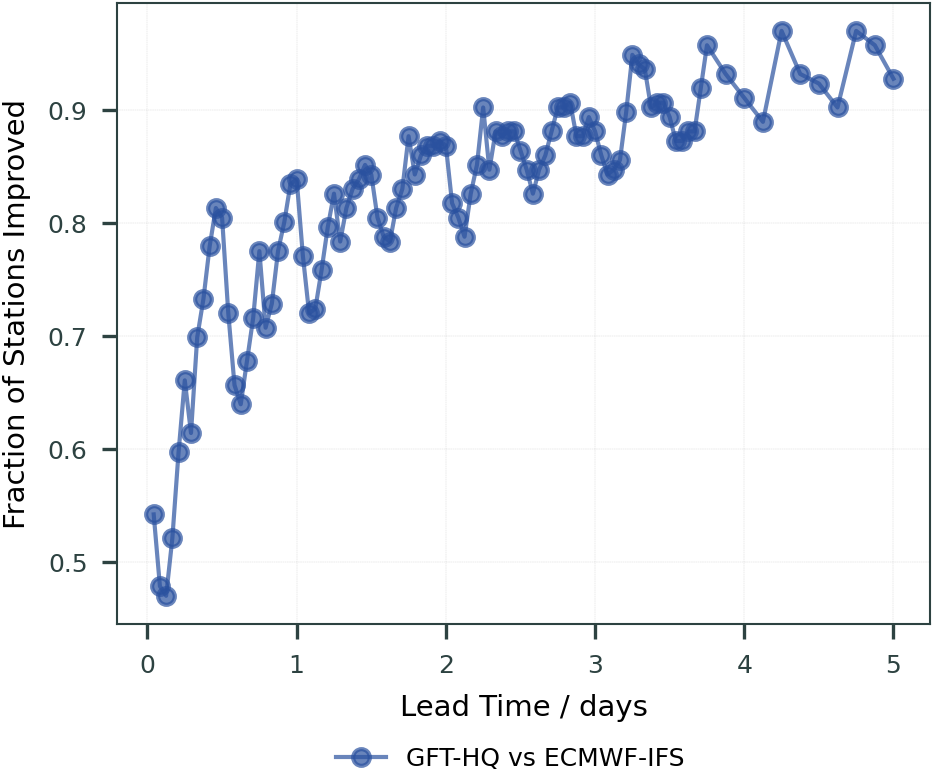}
      \put(2,82){\footnotesize\bfseries(\thesubfigure)}
    \end{overpic}
  \end{subfigure}
  \begin{subfigure}[b]{0.48\textwidth}
    \phantomsubcaption\label{fig:snap_temperature_evening}
    \begin{overpic}[width=\textwidth]{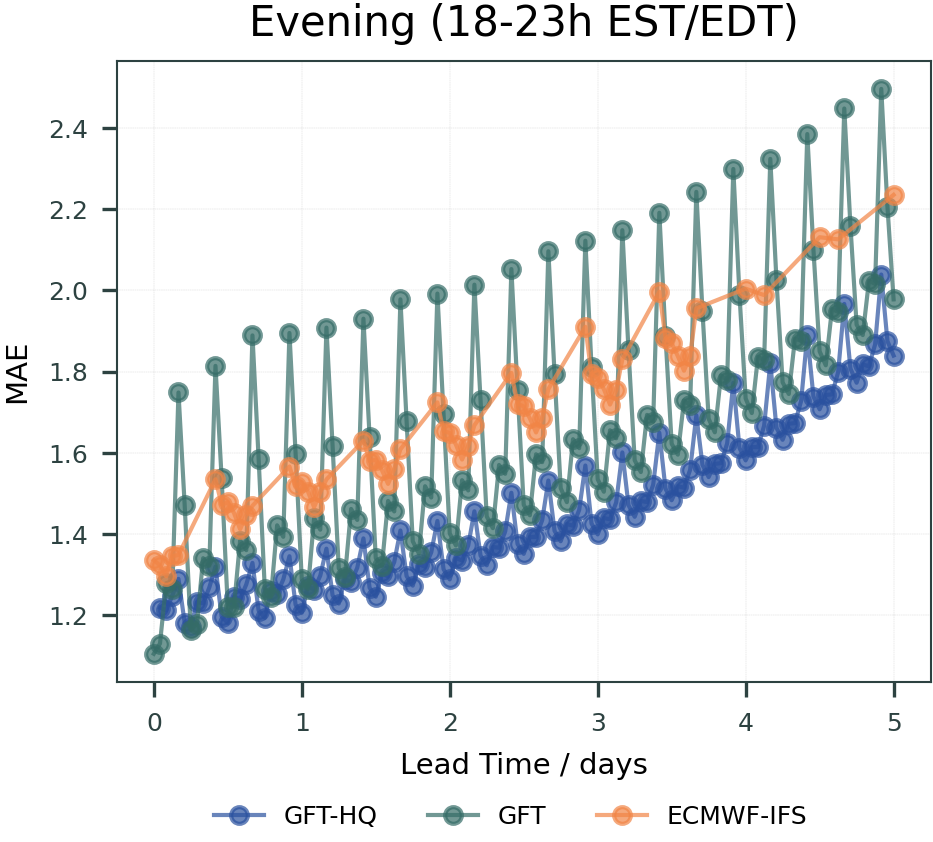}
      \put(2,82){\footnotesize\bfseries(\thesubfigure)}
    \end{overpic}
  \end{subfigure}
  \begin{subfigure}[b]{0.48\textwidth}
    \phantomsubcaption\label{fig:snap_temperature_summer}
    \begin{overpic}[width=\textwidth]{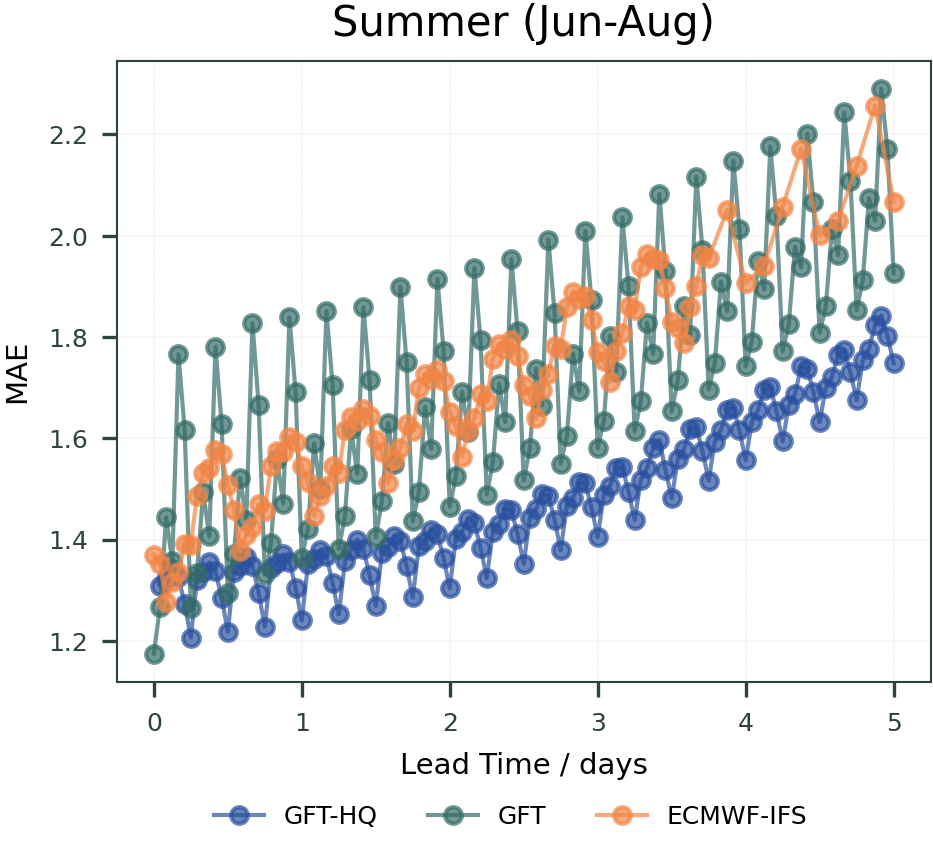}
      \put(2,82){\footnotesize\bfseries(\thesubfigure)}
    \end{overpic}
  \end{subfigure}
  \caption{Temperature forecast improvements from fine-tuning. (a) Forecast errors across all stations (b) Fraction of stations with lower forecasting errors than ECMWF-IFS (c) Mean absolute error across all stations in the evening during peak load (d) Mean absolute error across all stations in the unusually hot 2024 summer}
  \label{fig:temperature_main}
\end{figure}

\paragraph{Spatial footprint of gains.} Station-wise skill patterns (\Cref{app:temp-precip-forecasting}) show that post-training amplifies rather than redistributes the model’s strengths: GFT-HQ inherits the spatial structure of the untuned GFT but pushes most stations into the positive-skill regime by 24--48 h. \Cref{fig:temp_precip_maps_main} highlights the expanded positive-skill footprint for both temperature and precipitation. The broad coverage means load-forecast errors decline system-wide instead of concentrating at a few sites, and precipitation improvements propagate across the river basins that drive Hydro-Québec's reservoir operations. Diurnal and seasonal breakouts in \Cref{fig:temperature_main} (additional panels in \Cref{app:temp-precip-forecasting}) confirm that the gains hold during evening peaks and the extreme winter months that dominate risk planning.

\begin{figure}[t]
\begin{subfigure}[b]{0.48\textwidth}
    \phantomsubcaption\label{fig:snap_tp_mae} %
    \begin{overpic}[width=\textwidth]{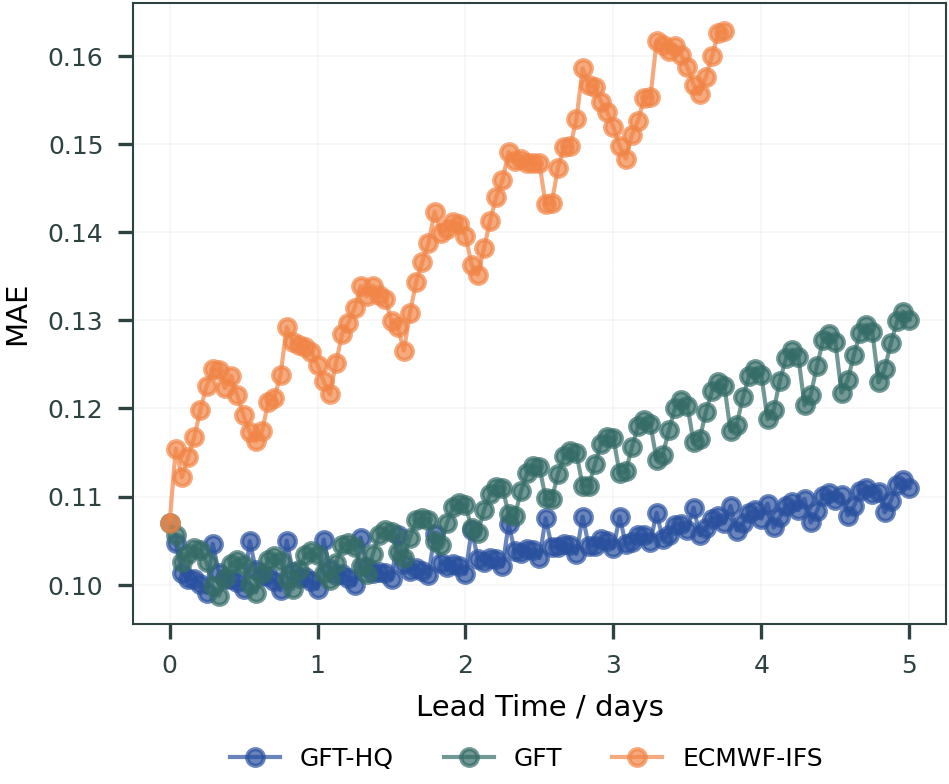}
      \put(2,82){\footnotesize\bfseries(\thesubfigure)}
    \end{overpic}
  \end{subfigure}
  \begin{subfigure}[b]{0.48\textwidth}
    \phantomsubcaption\label{fig:snap_tp_stations_improved}
    \begin{overpic}[width=\textwidth]{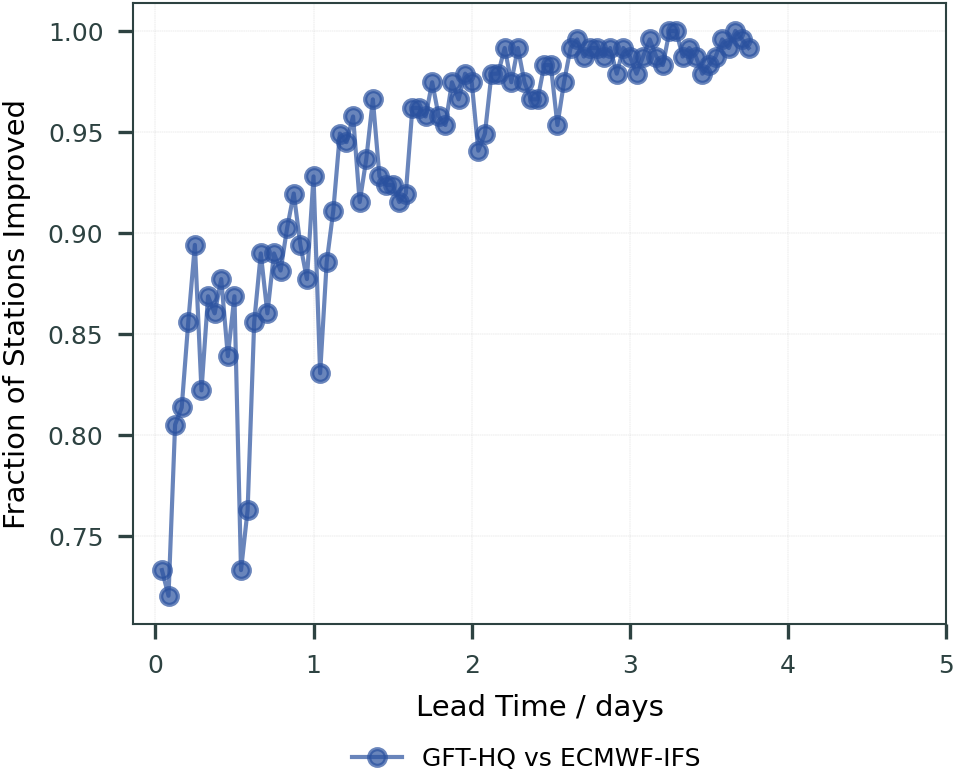}
      \put(2,82){\footnotesize\bfseries(\thesubfigure)}
    \end{overpic}
  \end{subfigure}
  \caption{Hourly total precipitation (mm/h) forecast improvements from fine-tuning. (a) Forecast errors across all stations (b) Fraction of stations with lower forecasting errors than ECMWF-IFS}
\end{figure}

\paragraph{Data}
We use hourly observations from federal/provincial networks and RMCQ  (534 temperature stations and 241 automatically validated precipitation stations, each with $\geq5$ years of data). We fine-tune the model on data from 2016-2023 and test on data from January 2024 to March 2025. See \Cref{app:datasets} for details.

\paragraph{Results}
\Cref{fig:snap_temperature_mae}  shows that GFT-HQ is consistently the lowest-error model for temperature forecasts across the entire 0–5-day horizon. Errors grow with lead time for all models, but the finetuned GFT-HQ model maintains a clear MAE gap to both GFT and ECMWF-IFS at every lead. The improvement is broadly distributed across sites: the fraction of stations with lower error than IFS rises from roughly a majority at short leads to $\approx 90-95\%$ by days 4-5. Gains are operationally aligned with HQ priorities: they are largest during evening peak hours (18-23 EST/EDT), and they persist through the anomalously hot 2024 summer, indicating robustness under regime shifts. In practical terms, trimming even a few-tenths of a degree in MAE at these horizons materially reduces load-forecast error and the reserve over-commit or imbalance penalties that follow.

\Cref{fig:snap_tp_mae} shows that GFT-HQ has the lowest MAE for precipitation forecasts at every lead from 0–5 days ($\approx 10\%$ better than GFT, $\approx 35\%$ better than ECMWF-IFS). The improvement is broadly distributed across sites: the fraction of stations beating IFS climbs to $\approx 90\%$ for day-ahead forecasts. Operationally, even a few-tenths of a tenth of a millimeter per hour compounds into meaningful 24-h accumulation error reductions, tightening day-ahead hydro-inflow and storm staffing plans.

\section{Conclusion}
Weather foundation models can materially improve weather intelligence for grid operations when post-trained on high-fidelity utility data. By adapting a pre-trained transformer based weather foundation model to Hydro-Québec assets, we produced hyper-local forecasts that outperformed strong NWP baselines on standard weather variables (temperature, precipitation, and wind). For rare but operationally critical hazards like rime ice, the model provides a new capability: reliable, hours-ahead alerts with usable precision-recall trade-offs. These gains translate into earlier and better-targeted interventions (e.g., de-icing; see \Cref{sec:deicing-decision-transmission}), more reliable renewable dispatch, and improved situational awareness at the asset level.

Looking ahead, we see three priorities for deployment: (i) decision-driven evaluation with utility-calibrated cost-loss parameters and crew constraints; (ii) tighter coupling with grid models (e.g., dynamic line ratings and outage risk) to convert forecast probabilities into asset risk; and (iii) continuous post-training and recalibration as new sensors come online. Beyond Québec, the same approach is directly transferable to other regions~\citep{silurian_gft_us_announce} and infrastructure owners, requiring only modest amounts of local data to adapt the model.

\subsubsection*{Acknowledgments}
We wish to thank Ana Poenariu, Guillem Candille, Alexis Trottier-Paquet, Alexandre Vidal, Pierre-Olivier Caron-Perigny, Sara-Ann Piscopo, Solutions Mesonet, Eric Morissette, Melanie Otis, Daniel Beland, Jean-Sebastien Ducharme-Aubé for their support.

\bibliography{tmlr}
\bibliographystyle{tmlr}

\appendix
\section{Appendix}

\subsection{Methods}
\label{app:methods}

\subsubsection{WFM post-training vs. NWP post-processing}
\label{app:wfm-vs-nwp-postproc}
This subsection expands the discussion in \Cref{sec:wfm-grid} by detailing how we adapt a pretrained weather foundation model (WFM), namely GFT, to Hydro-Québec assets via \emph{post-training} on utility observations. Conceptually, let $f_\theta$ denote the pretrained forecaster producing multi-variable fields and site forecasts at hourly lead times. Post-training updates $\theta$ using supervised losses defined on asset-level targets, so that the model's latent dynamics and decoders jointly produce hyper-local predictions. This approach fundamentally differs from traditional Numerical Weather Prediction (NWP) \emph{post-processing}.

The primary distinction lies in where the learning occurs. Conventional post-processing methods—such as Model Output Statistics (MOS), Ensemble Model Output Statistics (EMOS), or various machine learning models~\citep{glahn1972mos,gneiting2005emos,raftery2005bma,koenker1978qr,dellemonache2013anen}---learn a statistical mapping from a \emph{fixed} upstream NWP model's output to local observations. The NWP model itself remains unchanged. In contrast, our post-training approach directly updates the core forecasting model, $f_\theta$, enabling its internal spatiotemporal representations to better capture local physical phenomena like terrain and human infrastructure induced effects like icing.

This distinction leads to several key advantages:
\begin{itemize}
    \item \textbf{Multivariate Coherence:} Post-processing is typically performed on a per-variable, per-site basis, making it challenging to maintain physical and statistical consistency across different weather variables and lead times. While methods like the Schaake Shuffle~\citep{clark2004schaake} or Ensemble Copula Coupling (ECC)~\citep{schefzik2013ecc} can reconstruct some multivariate dependencies, our post-training framework is inherently a multi-task learning problem. By jointly predicting variables such as temperature, wind, precipitation, and icing, the model learns their interdependencies, ensuring greater physical coherence.
    \item \textbf{Generation of Novel Variables:} Post-processing is constrained by the output variables of the upstream NWP model. It cannot generate forecasts for phenomena not explicitly predicted by the NWP system. Our WFM, however, can be trained to predict new variables (here e.g., probability of rime ice on transmission lines) directly from local observations, leveraging the rich, shared latent state of the foundation model.
    \item \textbf{Scalability and Operational Value:} A traditional post-processing approach often requires developing and maintaining a large portfolio of individual models for each site and variable. The post-trained WFM provides a more scalable solution, with a single set of model weights serving hundreds of assets and multiple forecasting tasks. Because the fine-tuning adjusts the model's core dynamics rather than merely correcting surface-level biases, it yields more significant improvements in forecasting rare, high-impact events that are critical for operational decision-making.
\end{itemize}

\subsubsection{Experimental design}
\label{app:expdesign}
\textbf{Model Inputs and Targets}. At inference time, the encoder is initialized with dense ECMWF-IFS Analysis fields at the cycle issue time. The model is fine-tuned to decode to five targets used in Hydro-Québec operations: 2 m temperature, hourly precipitation, hub-height wind-speed, wind-farm icing indicator, and transmission-line rime-ice indicator. Unless explicitly stated, Hydro-Québec asset streams are used for supervision only and are not assimilated at runtime.

\textbf{Supervision and Data}. We employ standard regression losses for the continuous targets (temperature, wind, precipitation) and probabilistic classification losses for the binary icing indicators. The model is trained on observational data spanning from 2016 to 2023 and evaluated on a hold-out test set covering 2024 and 2025. Forecasts are evaluated at hourly cadence and also aggregated into decision windows (e.g., any icing in next 24 h) for operational metrics.

\textbf{Inference}. The post-trained model produces both dense gridded outputs and sparse hyper-local site forecasts in a single forward pass.  This unified workflow replaces the complex, multi-stage process of running an NWP model followed by a large suite of separate post-processing models.

\subsubsection{Evaluation metrics and interpretation}
\label{app:metrics}
We summarize the scores used in this work and their operational meaning for AI-based weather forecasting, building on the protocol introduced in the main text.

\paragraph{Base rate (class imbalance).} For a binary event indicator $y\in\{0,1\}$, the \emph{base rate} is $\pi = \tfrac{1}{N}\sum y$, i.e., the fraction of positive hours (or windows). Rime-ice at Sygivre sites is rare (example: $\pi\approx3.68\%$); wind-farm icing windows can be more frequent (example: $\pi\approx13\%$). Low $\pi$ makes precision--recall metrics more informative than accuracy.

\paragraph{MAE (mean absolute error).} For continuous variables (temperature, wind, precipitation), \emph{MAE}$=\tfrac{1}{N}\sum\lvert\hat y - y\rvert$ measures average magnitude of errors in native units (K, m/s, mm/h). Lower is better; it is robust and directly maps to operational tolerances (e.g., thermal ratings~\citep{ieee738_2013}, curtailment thresholds).

\paragraph{Precision, Recall, F1.} For thresholded probabilistic alerts $\hat y_\tau=\mathbb{1}[p\ge\tau]$:
\begin{align*}
\text{Precision} &= \tfrac{\mathrm{TP}}{\mathrm{TP}+\mathrm{FP}}, & \text{(false-alarm burden)}\\
\text{Recall (POD/TPR)} &= \tfrac{\mathrm{TP}}{\mathrm{TP}+\mathrm{FN}}, & \text{(missed-event rate)}\\
\text{F1} &= \tfrac{2\,\text{Precision}\cdot\text{Recall}}{\text{Precision}+\text{Recall}}. & \text{(balance under imbalance)}
\end{align*}
In operations, precision captures costly false dispatches; recall captures avoided misses; F1 summarizes the trade-off for rare hazards like icing.

\paragraph{PR curves, AP (PR-AUC), and lift.} Sweeping the alert threshold yields a precision--recall (PR) curve. \emph{Average Precision} (AP) is the area under the PR curve (also referred to as PR-AUC). AP increases when high-probability hours correspond to observed events. \emph{Lift} contextualizes AP under imbalance: $\text{lift}=\tfrac{\text{AP}}{\pi}$; a lift of $7\times$ means a random hour drawn from the model's top-ranked alerts is seven times likelier to be an event than an arbitrary hour.

\paragraph{ROC and AUC.} The ROC curve plots TPR vs FPR as $\tau$ varies; its area (AUC) lies in $[0,1]$. ROC-AUC is threshold-free and widely used, but with very low base rates it can overstate usefulness; we therefore emphasize PR/AP for icing and report ROC-AUC for completeness (e.g., wind-farm icing).

\paragraph{CSI (Critical Success Index) and IoU.} CSI (a.k.a. \emph{threat score}) measures categorical event performance at a fixed threshold: $\mathrm{CSI}=\tfrac{\mathrm{TP}}{\mathrm{TP}+\mathrm{FP}+\mathrm{FN}}$. In computer vision terms this is the \emph{Intersection over Union} (IoU) between the predicted and observed event sets; the two are equivalent for binary events in time or space.

\paragraph{Relative skill vs baseline.} When comparing to a baseline $b$ on an error metric $E$ (e.g., MAE) we report fractional skill
\[
\text{Skill} = 1 - \frac{E_{\text{model}}}{E_{b}}\,\,\in(-\infty,1],
\]
so that positive values indicate improvement (``positive = lower error than IFS'' in our maps). %

\paragraph{Windowed events.} For “any icing in the next $\Delta$ h” decisions we aggregate hourly probabilities $p_{t+h}$ into a window probability $q_t=1-\prod_h(1-p_{t+h})$ and then evaluate the same metrics on the windowed binary events; see \emph{Windowed event probability} in \Cref{sec:deicing-decision-transmission}.

\subsubsection{ERA5-derived Makkonen index}
\label{app:makkonen}
We construct a simple, physically motivated reference for rime-icing risk from ERA5 reanalysis fields (hourly, $0.25^{\circ}$ grid; \citealp{hersbach2020era5}). This ``Makkonen'' index \citep{makkonen2000models} is used only as a retrospective, physics-based benchmark in our comparisons and not as an operational forecast.

\paragraph{Sites and interpolation.} We take the locations of Hydro-Québec wind farms and Sygivre transmission-line sites and bilinearly interpolate ERA5 to each site at hourly cadence~\citep{hersbach2020era5}. Where site height above ground is unknown, we assign a fallback height of 80 m for wind farms and 50 m for transmission lines. To limit computation and reflect relevant icing layers, we use pressure levels from 800 to 1000 hPa and the time range 2024-01-01 to 2025-06-01.

\paragraph{Variables.} The following fields are used: 2 m temperature, 2 m dewpoint, 10 m and 100 m wind components (to form wind speed), surface pressure, surface geopotential, pressure-level geopotential, and specific cloud liquid water content on pressure levels from ERA5 at a site~\citep{hersbach2020era5}.
Given the hour, we derive:
\begin{itemize}
    \item \textbf{Wind speed at site height} $v(z)$ via a power-law profile using 10 m and 100 m winds:
    \begin{equation}
        v(z) = v_{10}\,\Big(\tfrac{z}{10}\Big)^{\alpha},\quad \alpha = \frac{\ln(v_{100}/v_{10})}{\ln(100/10)}.
    \end{equation}
    \item \textbf{Air temperature at site height} using a standard environmental lapse rate of 6.5 K km$^{-1}$ from 2 m: $T(z) = T_{2\,\mathrm{m}} - 6.5\,\mathrm{K\,km}^{-1}\,\tfrac{z}{1000}$.
    \item \textbf{Liquid water content (LWC) at site height} from the pressure-level specific cloud liquid water content, sampled at the site height using the geopotential field and converted to volumetric units (kg m$^{-3}$).
\end{itemize}

\paragraph{Icing and proxy rate.}
Following operational heuristics, rime icing is considered \emph{feasible} when all of the following hold:
\begin{equation}
    v(z) > 0\ \mathrm{m\,s}^{-1},\quad T(z) \in [260, 275]\ \mathrm{K},\quad \mathrm{LWC}(z) > 0.001\ \mathrm{g\,m}^{-3}.
\end{equation}
We then define a simple \emph{rate-of-icing proxy}
\begin{equation}
    r(t) = \begin{cases}
      v(z)\,\mathrm{LWC}(z), & \text{if icing feasible},\\
      0, & \text{otherwise},
    \end{cases}
\end{equation}
which intentionally omits geometry- and material-dependent collection efficiencies and thus serves only as a relative index. For 24 h events, we aggregate $r(t)$ over the decision window and threshold to produce a binary ``icing in next 24 h'' signal used in our comparisons.

\subsection{Datasets}
\label{app:datasets}
\subsubsection{Sygivre icing sensors}
\label{app:datasets-sygivre}
The Sygivre network comprises 40 atmospheric icing measurement stations, designed to measure either freezing rain or rime-ice accumulation in the vicinity of electricity transmission infrastructure. The main instrument at the stations is an ice-accretion measurement system. The central component of the system is a Rosemount 871 magnetorestrictive oscillation probe excited to vibrate at a natural frequency of 40 kHz \citep{kowles1973discussion}. As ice accretes on the probe, the inertia increases, and the corresponding decrease in vibration frequency is measured. While the Rosemount 871 is used widely to assess the occurrence of icing conditions, it is not meant for quantifying large accumulations of ice, which can encase the sensor and isolate it from the environment. For that reason, the second component of the system is a cylindrical mount for the probe, developed at the Hydro-Québec Research Center \citep{laforte1995wind}. The cylindrical mount contains a heater and an electromechanical system to shake the probe and remove ice completely, thus enabling continued measurements in large events. The final component is a controller that triggers the shaking and heating functions to rid the probe of ice at predetermined intervals. After measuring a decrease of 200 Hz, the probe heats up and shakes to remove the accumulated ice, and a ``de-icing cycle'' is logged. The power-line ice mass accretion corresponding to a 200 Hz decrease in probe frequency is calibrated separately in controlled experiments for dense freezing-rain ice and porous rime ice \citep{laforte1995wind}. An icing event is characterized by the number of ``de-icing cycles'' the probe undergoes, which provides an estimate of the ice mass accumulation. Of the 40 stations of the Sygivre network, 26 are in valley or coastal areas that are more exposed to freezing-rain events. The remaining 14 stations, used in the present study, are located at altitude near steep slopes and mostly record rime-ice events. Sygivre stations also include temperature, humidity, wind speed, and wind-direction sensors.

\subsubsection{Wind-farm meteorological masts}
\label{app:datasets-windmasts}
The wind data used here come from 65 out of a network of 70 permanent meteorological masts located near wind farms in Québec. The wind masts have been operational since wind-farm commissioning and have collectively recorded 800 mast-years of 10-minute-resolution data since 2006. Each mast includes at least three measurement levels between 10 and 130 meters. The masts are equipped with both heated and unheated sensors from manufacturers such as NRG Systems, Thies Clima, R.M. Young, Risø, WindSonic (Gill Instruments), and Vaisala. These sensors measure wind speed and direction, temperature, humidity, and atmospheric pressure. Data undergo automated quality control to remove invalid or erroneous values, though some uncertainty remains due to the absence of manual inspection. When multiple sensors are present at the same height, their validated measurements are combined into a single representative value per level. Ice-loss production for a given wind farm is not directly measured, but inferred from its potential production considering the wind speed and the temperature observed over all its turbines. The potential production is evaluated using the wind-farm power curve, which accounts for non-atmospheric losses (electrical losses, wake effects, etc.). The ice-loss production fraction is then taken as the ratio of the observed production of the wind farm to its potential production.

\subsubsection{Temperature and precipitation stations}
\label{app:datasets-rmcq}
The weather data used for post-training and evaluation of precipitation and temperature forecasts is gathered by the RMCQ (Quebec Collaborative Weather Network) as well as by Environment and Climate Change Canada and the Ministry of the Environment, the Fight Against Climate Change, Wildlife and Parks of Quebec. RMCQ centralizes weather station data shared by members, including utilities, mining companies and a wildfire prevention organization, that exploit weather stations spread over Quebec. The hourly temperature data used here are not validated and come from 534 stations, while the hourly precipitation data come from a subset of 241 stations and are automatically validated. Each station used in the study has at least the equivalent of 5 years of available data (43800 hours). Precipitation measuring stations are set up following WMO standards \citep{ISO19289,oke2018guide}. Precipitation data is validated automatically by the nonprofit Solutions Mesonet, using Oklahoma Climatological Survey guidelines. Validation includes steps like comparing nearest neighbours, flagging time discontinuities, thresholding for values that are unphysical or out of climatological range, and comparing similar instruments at a given station (e.g., Weighing Precipitation Gauge vs Tipping Bucket Rain Gauge).

\subsection{De-icing decision making}
\label{sec:deicing-decision-transmission}
Grid operators care less about generic skill and more about whether a forecast triggers the right crew action at the right time. We therefore evaluate the rime-ice head using a standard \emph{cost--loss} framework, adapted to partial loss avoidance by pre-emptive de-icing.

\paragraph{Setup.}
For each asset (or line segment) $i$ and time $t$, let $y_{t,i}\in\{0,1\}$ indicate whether at least one rime-ice accretion cycle occurs in the window $[t,t+\Delta]$ (e.g., $\Delta=24$ h). The model outputs a probability $p_{t,i}=\Pr(y_{t,i}=1\mid\mathcal{F}_t)$. The operator chooses $a_{t,i}\in\{\textsc{dispatch},\textsc{hold}\}$. Dispatching incurs a cost $C_{d,i}(t)$ (crew hours, travel, overtime), independent of $y_{t,i}$. If icing occurs and no dispatch is made, the operator faces a loss $L_i(t)$ (outage risk, repair, penalties). If icing occurs and dispatch happens, only a fraction $(1-\alpha_i(t))$ of this loss remains; $\alpha_i(t)\in[0,1]$ is the \emph{mitigation effectiveness} (fraction of loss avoided by pre-emptive action). Travel-time constraints $\tau_i(t)$ impose a lead-time requirement: dispatch is only effective if initiated before $t+\Delta-\tau_i(t)$.

\paragraph{Single-asset optimal threshold.}
Under these assumptions, the expected cost of \textsc{dispatch} is $\mathrm{EC}_{\textsc{dispatch}} = C_{d,i}(t) + p_{t,i}\,(1-\alpha_i(t))\,L_i(t)$ and the expected cost of \textsc{hold} is $\mathrm{EC}_{\textsc{hold}} = p_{t,i}\,L_i(t)$. \textsc{Dispatch} is optimal when $\mathrm{EC}_{\textsc{dispatch}}\le\mathrm{EC}_{\textsc{hold}}$, yielding the probability threshold $p^{\star}_{i}(t)$ reported in \Cref{eq:threshold}. Intuitively, crews mobilize when the icing probability exceeds the ratio “cost of acting” over “avoidable loss.”

\paragraph{Windowed event probability.}
Decisions target “any icing in the next $\Delta$ hours.” If $p_{t+h,i}$ is the per-hour probability for $h=0,\ldots,H-1$ with $H=\Delta/1\,\text{h}$, the event probability over the window is the union
\[
q_{t,i} \;=\; 1 - \prod_{h=0}^{H-1} \big(1-p_{t+h,i}\big).
\]
When using the window mean $\bar p_{t,i}=H^{-1}\sum_h p_{t+h,i}$ as a score, a Poisson/independence approximation yields $q_{t,i} \approx 1-\exp(-H\,\bar p_{t,i})$. The optimal threshold on $q$ maps to a threshold on $\bar p$ via
\[
\bar p^\star_{i}(t) \;=\; -\frac{\ln\big(1-p^\star_{i}(t)\big)}{H}.
\]

\paragraph{Lead-time aware policy.}
Let $\tilde{p}_{t,i}$ denote the forecast issued at time $t$ for the next $\Delta$ hours. Define the effective window as those issue times that satisfy the travel constraint $t\le t_{\text{event}}-\tau_i(t)$ for any $t_{\text{event}}\in[t,t+\Delta]$. We evaluate at a 6-hour issuance cadence (00, 06, 12, 18), trigger a \textsc{watch} when $\tilde{p}_{t,i}\ge \eta\,p^\star_i(t)$ and a \textsc{dispatch} when $\tilde{p}_{t,i}\ge p^\star_i(t)$, with $0<\eta<1$ (e.g., $\eta=0.7$) providing earlier mobilization. To avoid “thrashing,” we apply hysteresis with two thresholds $p^\star_{\text{on}}>p^\star_{\text{off}}$ and a persistence requirement across issuance steps (e.g., exceedance for $k=2$ consecutive issuances).

\begin{figure}[t]
    \centering
    \includegraphics[width=0.99\linewidth]{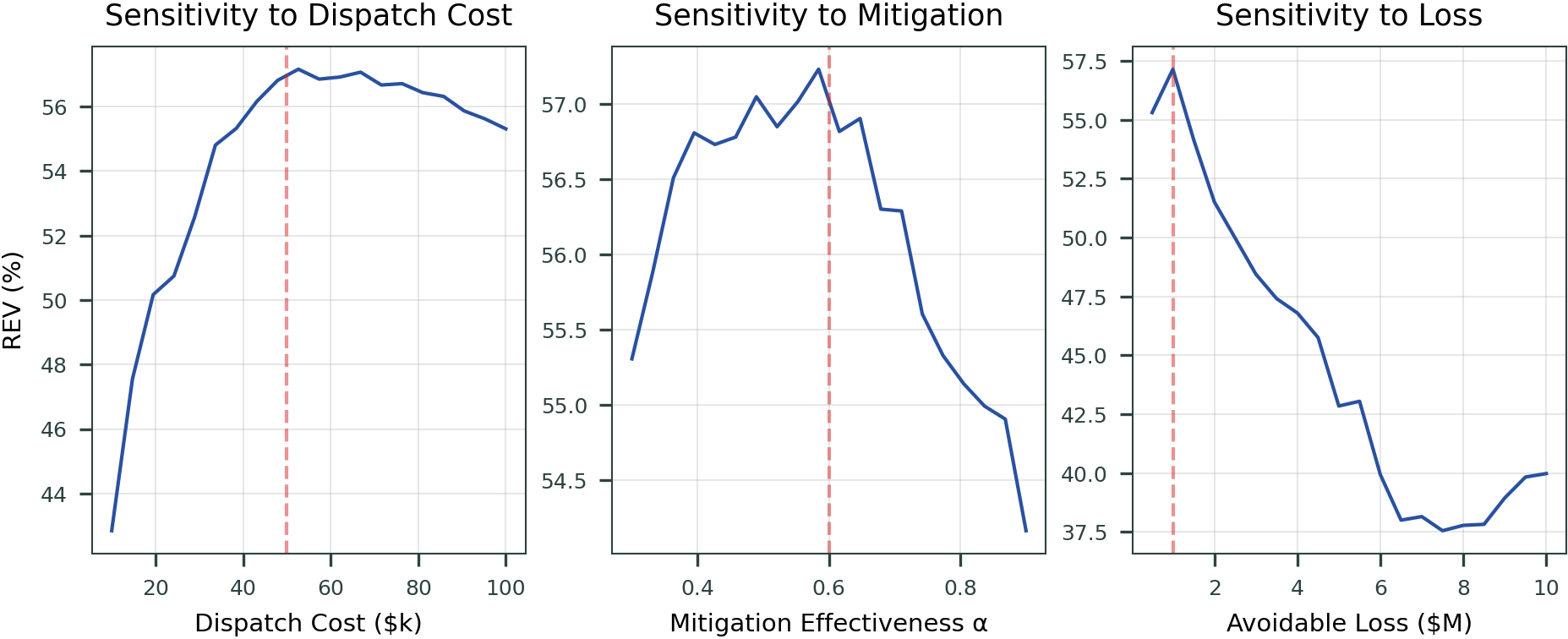}
    \caption{REV sensitivity: one-at-a-time variation of dispatch cost $C_d$, mitigation effectiveness $\alpha$, and avoidable loss $L$; vertical dashed lines mark the base setting.}
    \label{fig:rev_sensitivity}
\end{figure}

\begin{table}[t]
\centering
\caption{Default de-icing decision parameters and example thresholds.}
\begin{tabular}{lcc}
\toprule
Parameter & Default \\
\midrule
Decision window $\Delta$ & 24 h \\
Issuance cadence & 6 h  \\
Lead-time (mobilization) $\tau$ & 0 h  \\
Mitigation effectiveness $\alpha$ & 0.6  \\
Crew capacity $H(t)$ & 3 crews  \\
Per-asset crew-hours $h_i(t)$ & 12 h \\
Dispatch cost $C_d$ (ground) & \$50k  \\
Helicopter adder & \$75k \\
Avoidable loss $L$ & \$400k  \\
Hysteresis/persistence & $p_\text{on}=p^\star$, $p_\text{off}=0.6\,p^\star$, $k=2$ \\
\bottomrule
\end{tabular}
\end{table}

\paragraph{Scope and limitations.}
The dollar values and parameter ranges used here are \emph{illustrative} and chosen to make the framework concrete rather than to prescribe operational thresholds. 
Real systems exhibit wide variability by terrain and access (helicopter vs.\ ground), voltage class, outage externalities, work rules, and weather. 
We also abstract from several complexities: (i) heavy–tailed and network–coupled losses (e.g., cascading redispatch and penalties), (ii) routing, staging, backlog and queueing across multi–day events, (iii) shared crews and switching clearances.

\paragraph{Measuring decision value.}
We report the \emph{relative economic value} (REV) of the forecast policy against both a climatology policy and a perfect-forecast oracle:
\begin{equation}
\mathrm{REV} = \frac{\mathbb{E}[C_{\text{clim}}]-\mathbb{E}[C_{\text{fcst}}]}{\mathbb{E}[C_{\text{clim}}]-\mathbb{E}[C_{\text{perf}}]}\ \in[0,1],
\end{equation}
where expected costs are averaged over the held-out period. Here $C_{\text{fcst}}$ is the realized cost under the threshold/hysteresis policy above, $C_{\text{clim}}$ is an instance-wise climatology baseline obtained by applying the same decision rule to climatology probabilities $C_{\text{perf}}$ assumes perfect knowledge of $y_{t,i}$.

We include a one-way sensitivity analysis to $(C_d,\alpha,L)$ to show robustness of REV and the optimal threshold.

\paragraph{Base scenario results.}
With $C_d=\$50{,}000$, $\alpha=0.6$, and $L=\$400{,}000$ (yielding $p^\star\approx20.8\%$ on the 24 h event probability), we obtain:
\begin{itemize}
    \item $\text{REV}=57.2\%$;
    \item $C_{\text{fcst}}=\$51{,}913$; $C_{\text{clim}}=\$69{,}064$; $C_{\text{perf}}=\$39{,}055$;
    \item base rate $\mathbb{E}[y]=8.7\%$.
\end{itemize}
Thus the forecast reduces expected cost by \$17{,}151 relative to the climatology baseline and closes about 57\% of the gap to perfect information: $(69{,}064-51{,}913)/(69{,}064-39{,}055)\approx0.572$.

\subsection{Related Work}

\textbf{Traditional NWP vs. AI-based Weather Forecasting}: Weather forecasting has long been dominated by physics-based numerical weather prediction (NWP) models, which solve discretized fluid dynamics equations but at enormous computational cost. Operational global models like ECMWF's IFS require supercomputers and still make simplifying assumptions (e.g., subgrid parametrizations) that limit accuracy. In recent years, data-driven systems have advanced rapidly, offering orders-of-magnitude faster inference with increasingly competitive skill. More recently, systems including DeepMind's GraphCast~\cite{graphcast2023} and ECMWF's AIFS~\cite{chantry2025aifs} demonstrated that learned surrogates can match or surpass operational guidance on many metrics at a fraction of the compute, catalyzing a shift toward hybrid and ML-first forecasting; see also \citep{bouallegue2024rise} for an operational-like statistical assessment. Short-range precipitation nowcasting has also seen strong ML gains~\citep{ravuri2021dgmr,sonderby2020metnet}. 

\textbf{Foundation Models for Weather and Climate}: Inspired by the paradigm of large-scale pretraining in NLP and vision, researchers began developing foundation models for Earth science that can be adapted to many downstream tasks~\cite{bodnar2025aurora,climax23,bi2023pangu,pathak2022fourcastnet,chen2023fuxi,szwarcman2024prithvi}. Earliest example was ClimaX, a deep learning model trained on heterogeneous weather/climate datasets in 2023. ClimaX demonstrated that a single pre-trained model could be fine-tuned to diverse tasks, from short-range forecasts of standard variables to climate projections and downscaling. More importantly, it often matched conventional specialized models on benchmark tests. More recently, Aurora pushed this concept further: a 1.3-billion-parameter transformer that unified multiple Earth-system domains within one model. After pretraining on over a million hours of geophysical data, Aurora was post-trained (fine-tuned) for specific high-value applications, achieving state-of-the-art accuracy in each. Notably, Aurora outperformed operational NWP systems in 10-day global weather forecasting (improving error on 92\% of targets vs. ECMWF's IFS), 5-day tropical cyclone tracking (20\% lower track error than the multi-agency consensus), 5-day air quality (beat ECMWF's chemistry model on 74\% of metrics), and 10-day ocean wave prediction (beat ECMWF's wave model on 86\% of metrics), all at a fraction of the computing cost. Crucially, these feats were achieved by fine-tuning the same pretrained model for only a few hours per task, without retraining the core architecture. This demonstrates the power of foundation models: once a model has “learned the grammar of the Earth system” in its latent representations, it can be rapidly adapted to new domains with limited data. The Generative Forecasting Transformer (GFT) family extends this idea even further. GFT surpasses Aurora's global weather skill on all metrics, add finer temporal resolution (hourly forecasts), and expand the set of predicted variables. A regional variant, GFT-US, now delivers kilometer-scale 0–30 h forecasts tailored to the United States. In summary, the field has rapidly evolved from single-purpose neural nets to general-purpose weather foundation models (WFMs) that can democratize high-quality forecasts by post-training on specific tasks at modest cost. Our work builds directly on these advances—particularly Aurora and its GFT successors—by exploring how a WFM can be leveraged in the power grid domain.

\textbf{NWP post-processing vs. WFM post-training}: Utilities have long relied on statistical post-processing to adapt NWP outputs to sites and assets. Common approaches include bias correction and calibration such as Model Output Statistics (MOS), ensemble MOS (EMOS), Bayesian model averaging (BMA), quantile-based methods, analog ensembles, and neural/ML regressors~\citep{glahn1972mos,gneiting2005emos,raftery2005bma,koenker1978qr,dellemonache2013anen,rasp2018nnpost,taillardat2016qrf,wilks2011statistical,vannitsem2018postprocessing}. These methods learn a mapping from fixed NWP fields to local targets, typically one variable at a time and one site at a time, and they do not alter the upstream atmospheric dynamics. By construction, they cannot introduce truly new target variables that are absent from the NWP (e.g., rime-ice risk) and often struggle to maintain cross-variable, cross-lead physical coherence; dependency-preserving techniques like the Schaake Shuffle and Ensemble Copula Coupling (ECC) address some multivariate aspects but do not modify the underlying flow predictions~\citep{clark2004schaake,schefzik2013ecc}.
In contrast, our approach adapts the forecasting \emph{model itself}. We fine-tune a pretrained WFM end-to-end on utility observations so that its latent dynamics and decoders jointly produce hyper-local, multi-variable predictions. This confers several advantages for operations: (i) multi-task training enforces consistency across variables (e.g., wind, temperature, precipitation, icing) and lead times; (ii) the model’s sparse hyper-local decoder targets specific assets without training a separate model per site; (iii) new targets like rime-ice probability can be learned directly from observations; and (iv) updates propagate into the model’s spatiotemporal representation, improving fine-scale phenomena rather than merely correcting biases at the output. Empirically, we observe that a single post-trained WFM can replace a portfolio of site-specific post-processing models while delivering higher skill, especially for rare, high-impact events.

\textbf{Weather Forecasting for Power Systems}: The electric power grid is highly weather-sensitive, with assets and operations affected by temperature, wind, precipitation, icing, etc. Traditionally, utilities have relied on NWP model outputs combined with statistical corrections for site-specific forecasting. For instance, wind power forecasting methods have long used physical NWP feeds to predict turbine wind speeds, often augmented by local regression models~\cite{yang2024survey}. Beyond generation forecasts, weather hazards pose a major threat to grid infrastructure. Icing of transmission lines is a prime example: accreted ice from freezing rain or rime can add massive weight and cause line sag or breakage, leading to catastrophic outages~\citep{makkonen2000models}. Icing also degrades wind-turbine production and availability in the field~\citep{gao2021field}. This has made accurate icing forecasts increasingly crucial for grid reliability and disaster preparedness.

Forecasting line icing is notoriously challenging. Many icing forecast methods build on numerical weather prediction (NWP) models to provide the meteorological inputs for ice accretion models. High-resolution regional models (e.g., WRF~\citep{skamarock2019wrf}) can simulate temperature, wind, humidity, and precipitation fields that drive icing formation. Physical ice accretion models use these weather parameters to estimate ice growth on structures based on thermodynamic and empirical relationships~\citep{makkonen2000models}. \citet{musilek2009ice} developed an Ice Accretion Forecasting System (IFAS) that ingests NWP model data (including a precipitation-type algorithm by Ramer) to predict freezing rain occurrence and ice loads on power lines. Data-driven approaches have also been explored for transmission-line icing prediction~\citep{li2014multivariable}.

However, purely physical models face limitations. NWP models have constrained resolution and tend to smooth out terrain effects, so they may miss localized microclimate conditions (e.g., rime icing on a ridge or heavier glaze in a valley). This mismatch means a weather model might predict the general area of an icing event but still underestimate or misplace the peak ice loads. \citet{wang2024prediction} used a WRF model with 3D variational data assimilation to simulate a power line icing event. Local weather observations improved the accuracy of temperature and humidity fields (especially in the first 24 hours), which in turn made the icing forecasts more realistic near instrumented sites. However, WRF models are significantly more expensive to tune and operate compared to WFMs.

\subsection{Additional Results}

\subsubsection{Rime ice forecasting for transmission lines}
\label{app:transmission-forecasting}

\begin{figure}
    \centering
    \begin{subfigure}[b]{0.48\textwidth}
    \includegraphics[width=\linewidth]{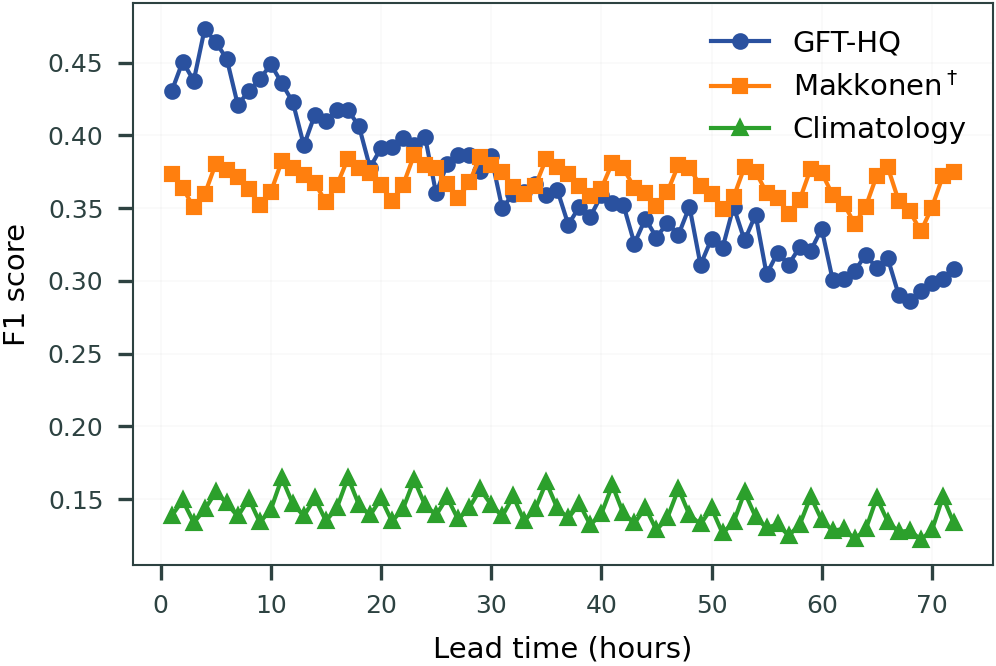}
    \caption{Hourly (1 h)}\label{fig:transmission_line_icing_hourly_f1_hourly}
    \end{subfigure}
    \begin{subfigure}[b]{0.48\textwidth}
        \includegraphics[width=\linewidth]{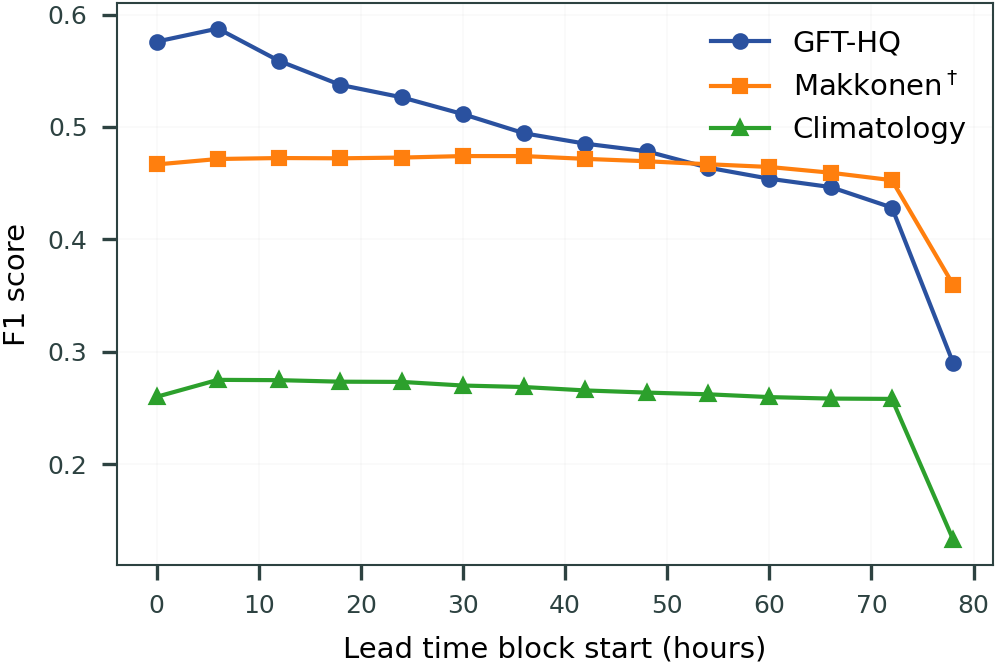}
        \caption{6 h windows}\label{fig:transmission_line_icing_hourly_f1_6h}
    \end{subfigure}
    \caption{Lead-time dependence of transmission-line rime-ice detection skill. Left (\textbf{a}): hourly F1 vs. lead time (1–72 h). Right (\textbf{b}): F1 for 6-hour windowed targets (``any icing in window'') vs. window start time. Aggregating to 6 h smooths timing errors and better reflects current HQ operations, producing higher and more stable scores. In both settings, GFT-HQ achieves the highest skill and is comparable to or better than the ERA5-derived Makkonen reference through 72 h; climatology trails substantially. All scores are pooled across the 14 Sygivre stations.}
    \label{fig:transmission_line_icing_hourly_f1}
\end{figure}

\Cref{fig:transmission_line_icing_hourly_f1_hourly} shows hour-by-hour F1. \Cref{fig:transmission_line_icing_hourly_f1_6h} evaluates 6-hour windows, which is closer to current HQ operations; this aggregation reduces penalties from small timing offsets and leads to higher, more stable scores across lead times while preserving the ranking between methods.

\subsubsection{Wind-farm wind speed and ice forecasting}
\label{app:windfarm-forecasting}

\begin{figure}[t]
    \centering
    \includegraphics[width=0.5\linewidth]{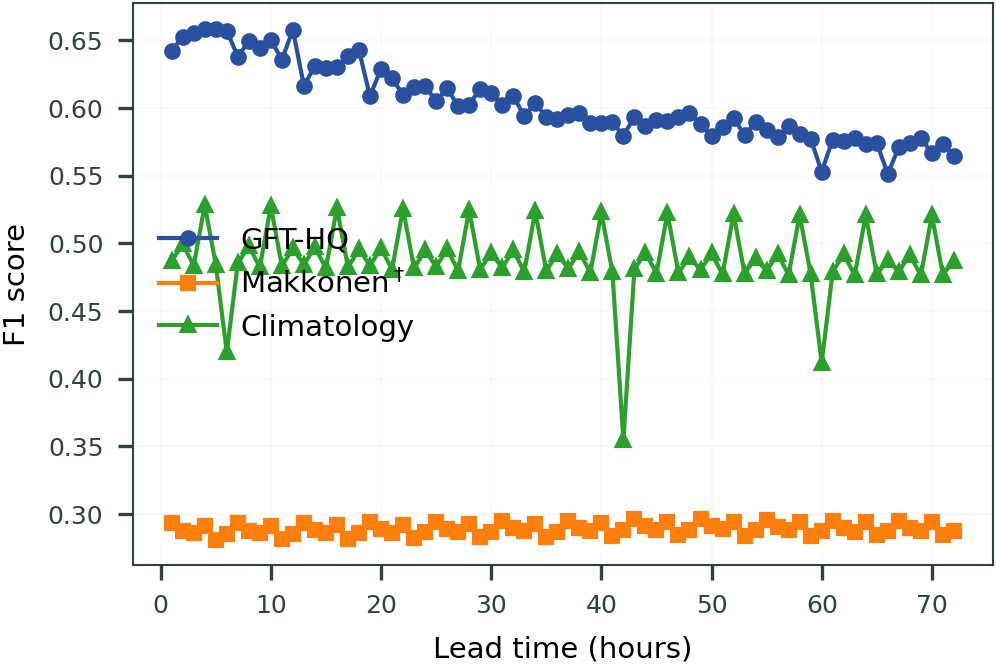}
    \caption{Lead-time dependence of wind-farm icing risk forecasts. F1 score vs. lead time (1-72h), pooled across all windfarms. GFT-HQ maintains the highest skill at all leads ($\approx 0.65$ at short leads, tapering to $\approx  0.58$ by 72 h).}
    \label{fig:windfarm_icing_hourly_f1}
\end{figure}

\paragraph{Consistent icing risk forecasts}
\Cref{fig:windfarm_icing_hourly_f1} compares the F1 score over different lead-times of the forecast with GFT-HQ maintaining high skill for more than 2-day lead times.

\begin{figure}[t]
    \begin{subfigure}[b]{0.48\textwidth}
        \centering
        \includegraphics[width=\textwidth]{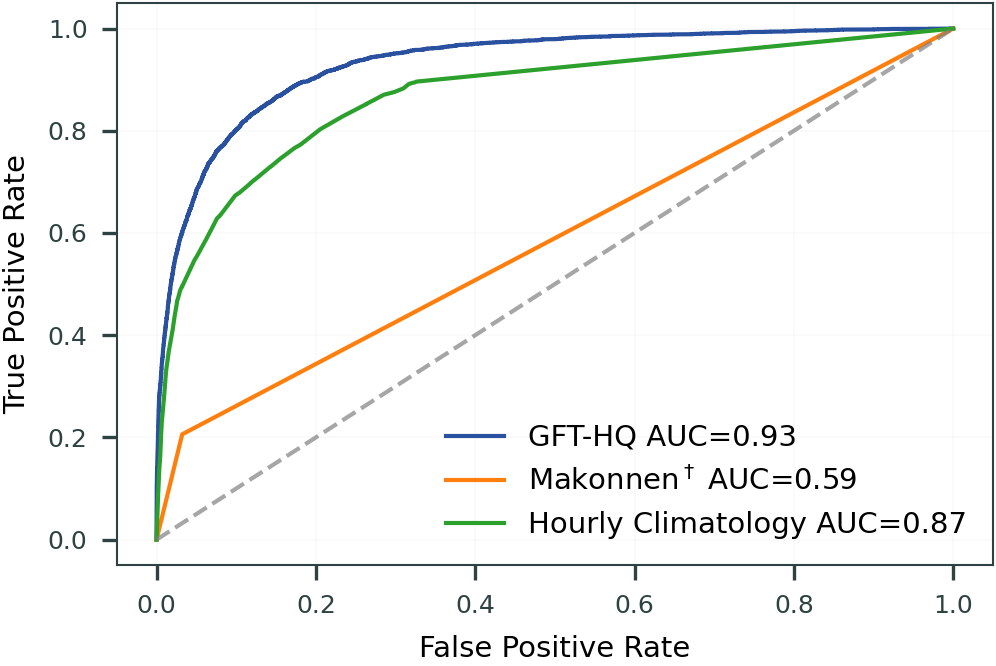}
        \caption{Forecast at 24h lead-time}
    \end{subfigure}
    \begin{subfigure}[b]{0.48\textwidth}
        \centering
        \includegraphics[width=\textwidth]{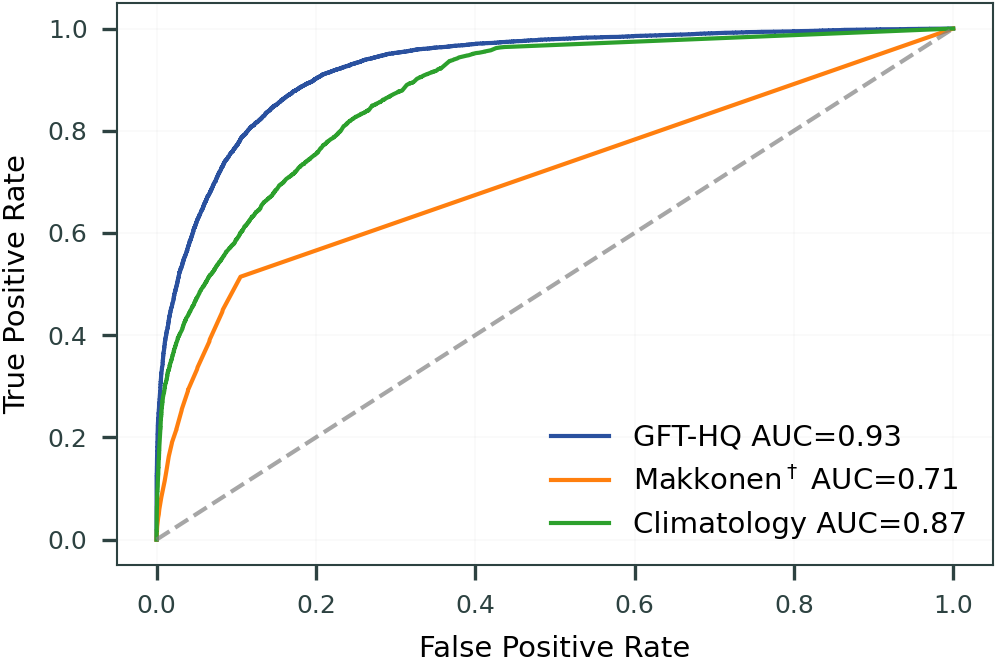}
        \caption{Forecast over 24h time-horizon}
    \end{subfigure}    
    \caption{Wind-farm icing ROC-AUC curves. }
    \label{fig:roc_auc_windfarm}
\end{figure}

As can be observed in \Cref{fig:roc_auc_windfarm}, corresponding ROC performance is high as well (AUC = 0.93 vs 0.86–0.87 (Climatology) and 0.59–0.71 (Makkonen) in the hourly and 24-hour windowed setups.

\begin{figure}[t]
\begin{subfigure}[b]{0.48\textwidth}
    \phantomsubcaption\label{fig:snap_ws_mae_morning} %
    \begin{overpic}[width=\textwidth]{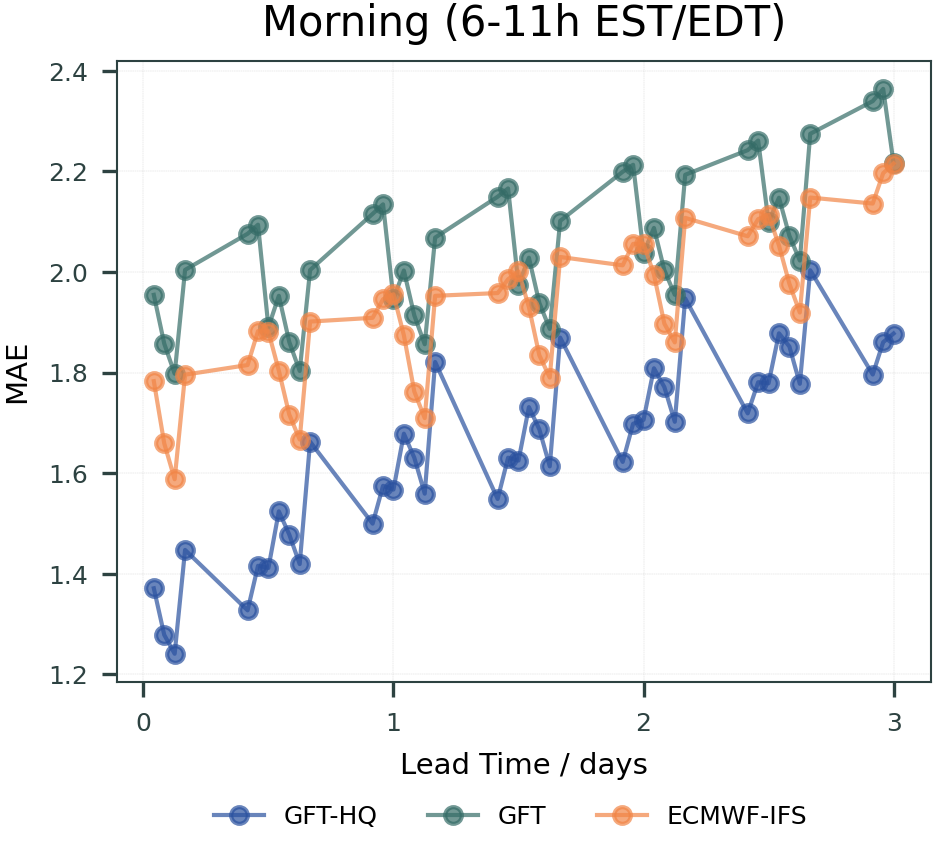}
      \put(2,82){\footnotesize\bfseries(\thesubfigure)}
    \end{overpic}
  \end{subfigure}
  \begin{subfigure}[b]{0.48\textwidth}
    \phantomsubcaption\label{fig:snap_ws_mae_afternoon}
    \begin{overpic}[width=\textwidth]{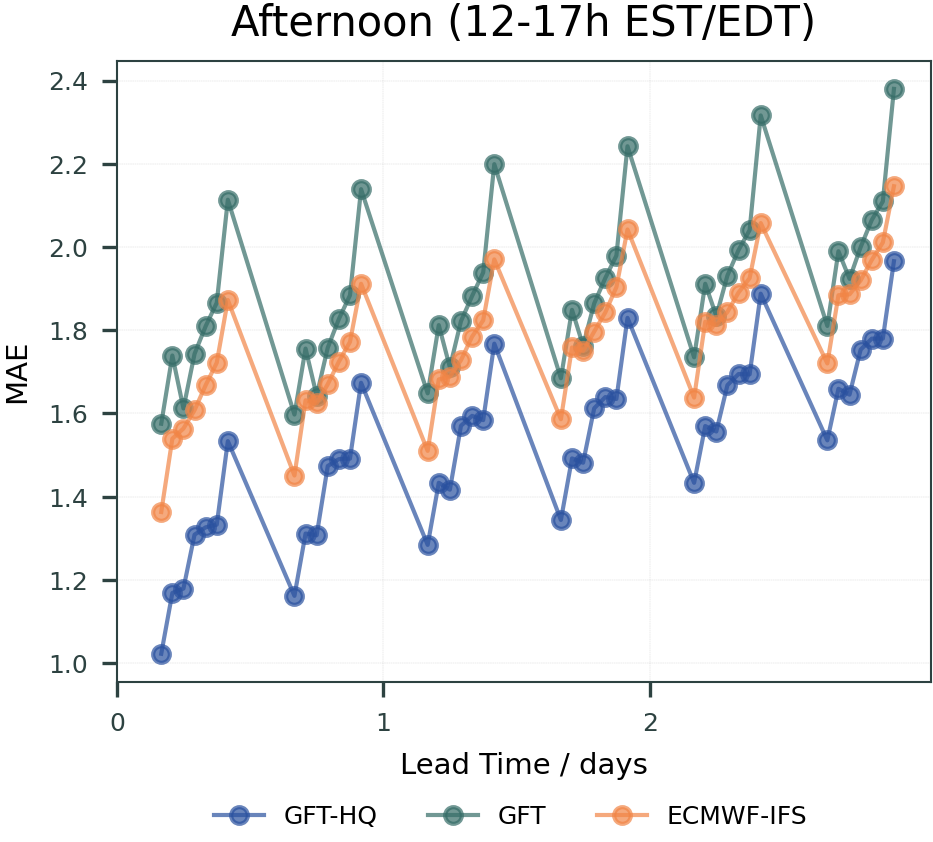}
      \put(2,82){\footnotesize\bfseries(\thesubfigure)}
    \end{overpic}
  \end{subfigure}
  \begin{subfigure}[b]{0.48\textwidth}
    \phantomsubcaption\label{fig:snap_ws_mae_evening} %
    \begin{overpic}[width=\textwidth]{figs/windfarm/windspeed_Evening_mae.png}
      \put(2,82){\footnotesize\bfseries(\thesubfigure)}
    \end{overpic}
  \end{subfigure}
  \begin{subfigure}[b]{0.48\textwidth}
    \phantomsubcaption\label{fig:snap_ws_mae_night}
    \begin{overpic}[width=\textwidth]{figs/windfarm/windspeed_Night_mae.png}
      \put(2,82){\footnotesize\bfseries(\thesubfigure)}
    \end{overpic}
  \end{subfigure}  
  \caption{Hourly hub-height wind speed forecast improvements from fine-tuning: Station-averaged MAE for \textcolor{blue}{GFT-HQ}, \textcolor{green!60!black}{GFT}, and \textcolor{orange}{ECMWF-IFS}, grouped by local time of the day. GFT-HQ is lowest-error across all panels and leads, with the largest improvements in \emph{Evening/Night} and a sustained gap through the day-ahead window. (MAE units are m s$^{-1}$; vertical oscillations reflect the diurnal verification hour across lead time. }
  \label{fig:wind_daypart}
\end{figure}

\begin{figure}[t]
\begin{subfigure}[b]{0.48\textwidth}
    \phantomsubcaption\label{fig:snap_ws_mae_spring} %
    \begin{overpic}[width=\textwidth]{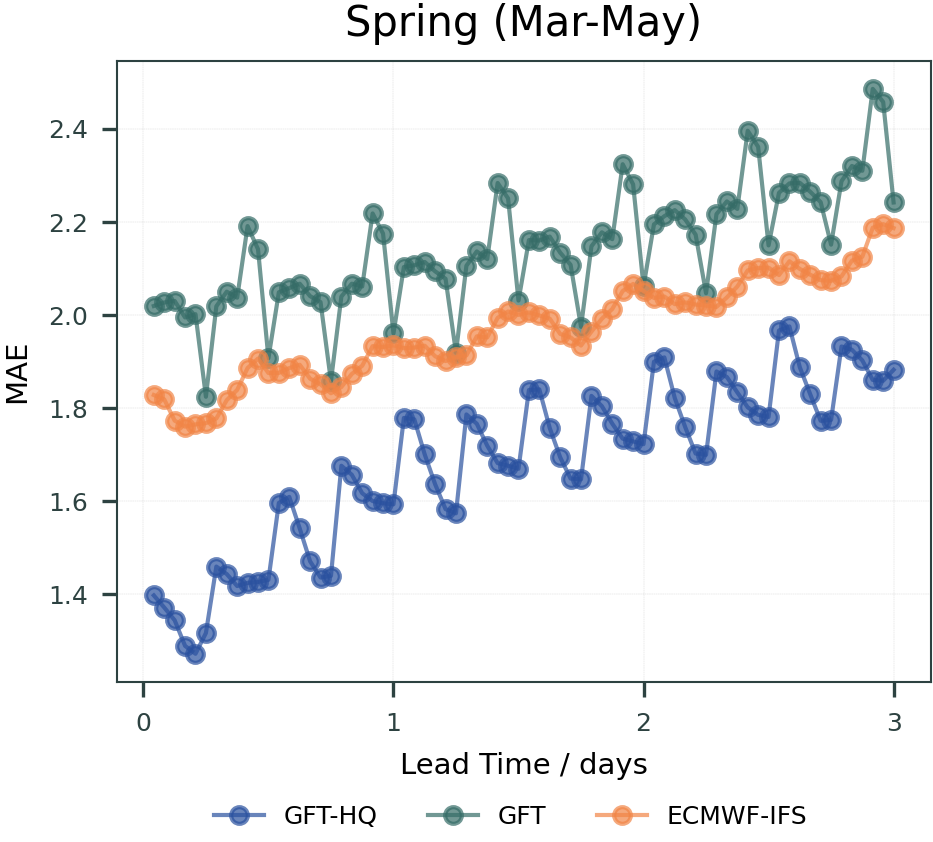}
      \put(2,82){\footnotesize\bfseries(\thesubfigure)}
    \end{overpic}
  \end{subfigure}
  \begin{subfigure}[b]{0.48\textwidth}
    \phantomsubcaption\label{fig:snap_ws_mae_summer}
    \begin{overpic}[width=\textwidth]{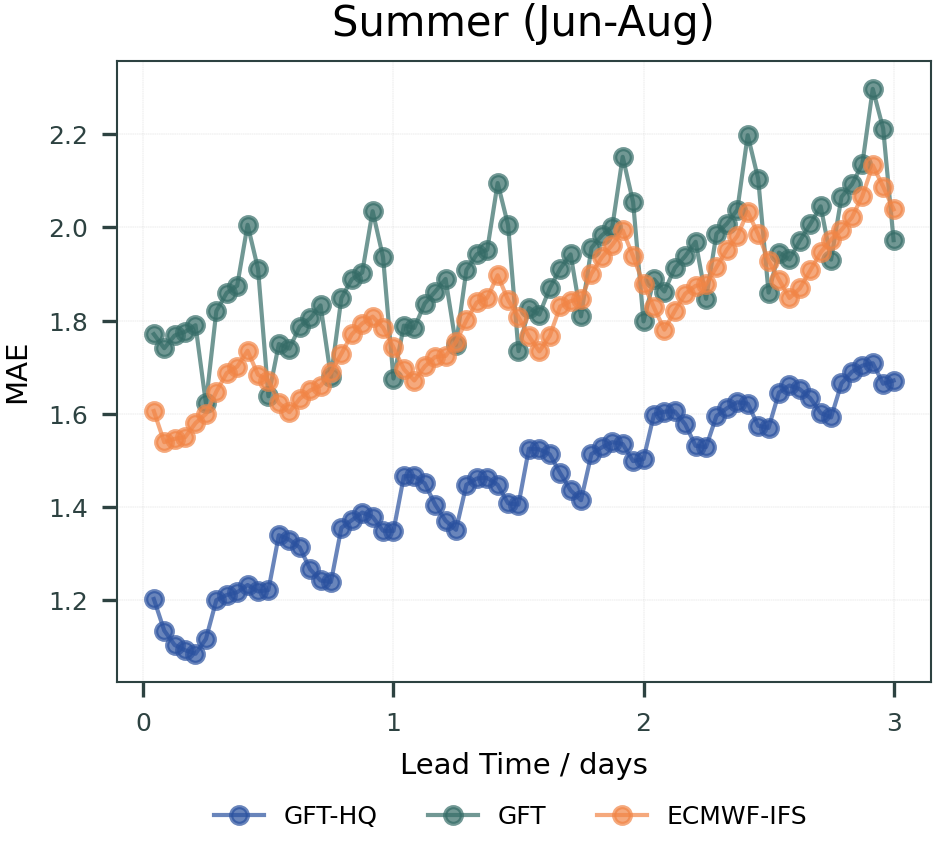}
      \put(2,82){\footnotesize\bfseries(\thesubfigure)}
    \end{overpic}
  \end{subfigure}
  \begin{subfigure}[b]{0.48\textwidth}
    \phantomsubcaption\label{fig:snap_ws_mae_fall} %
    \begin{overpic}[width=\textwidth]{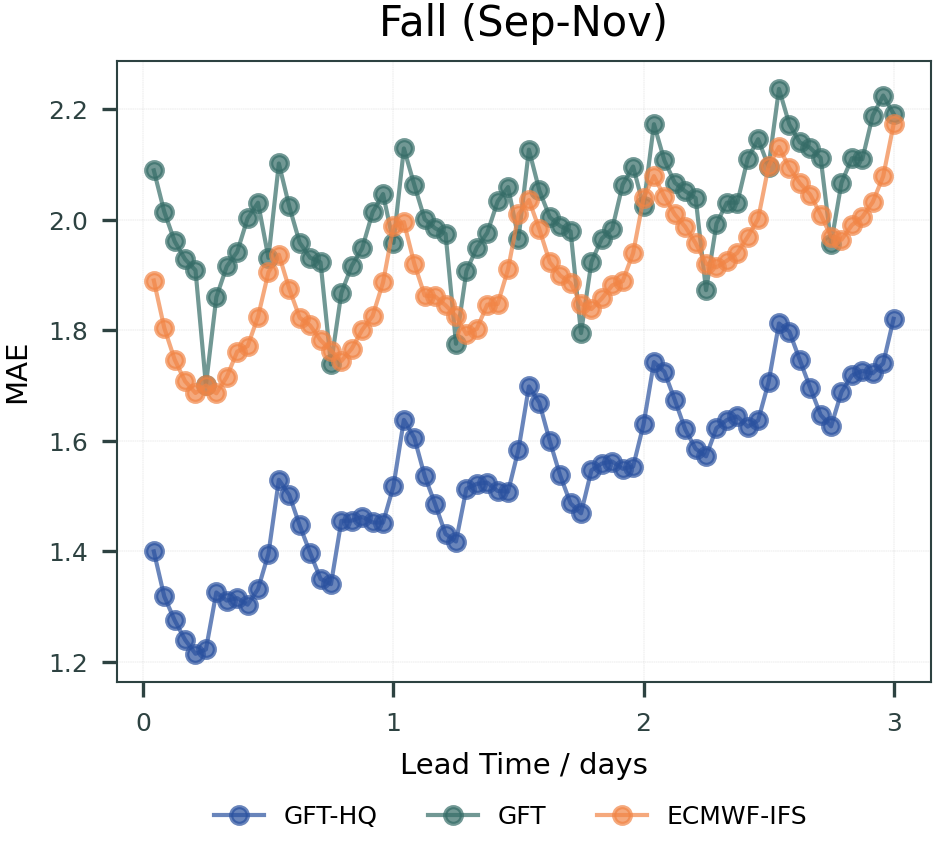}
      \put(2,82){\footnotesize\bfseries(\thesubfigure)}
    \end{overpic}
  \end{subfigure}
  \begin{subfigure}[b]{0.48\textwidth}
    \phantomsubcaption\label{fig:snap_ws_mae_winter}
    \begin{overpic}[width=\textwidth]{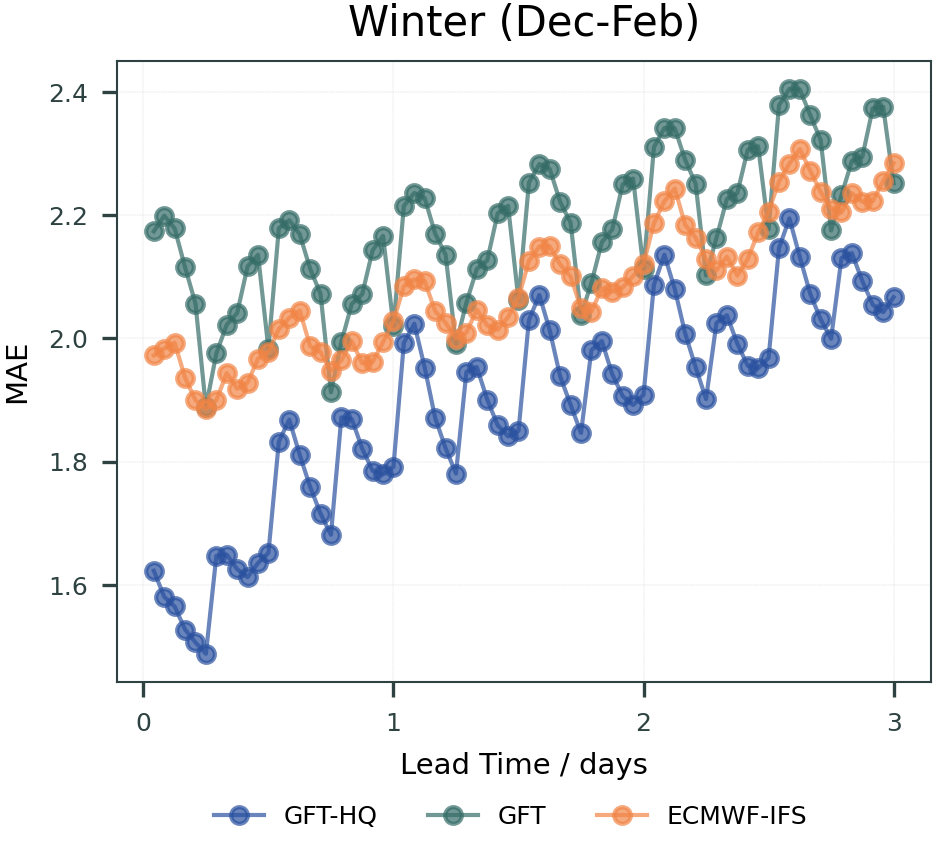}
      \put(2,82){\footnotesize\bfseries(\thesubfigure)}
    \end{overpic}
  \end{subfigure}  
  \caption{Hub-height wind speed MAE by season:  Station-averaged MAE for \textcolor{blue}{GFT-HQ}, \textcolor{green!60!black}{GFT}, and \textcolor{orange}{ECMWF-IFS}, grouped by season. GFT-HQ improves on skill in all seasons. However, the error patterns suggest that it could be helpful to oversample winter season to further improve performance in that time-period.}
  \label{fig:wind_season}
\end{figure}

\paragraph{Consistent Performance Across Diurnal and Seasonal Cycles.}
We analyzed the hub-height wind forecast skill of GFT-HQ against baseline models across different times of day (\Cref{fig:wind_daypart}) and seasons (\Cref{fig:wind_season}). The results demonstrate a robust and consistent performance advantage for the finetuned model.

A key finding is that GFT-HQ significantly dampens the error oscillations present in the baseline models. While the original GFT exhibits a pronounced diurnal error ripple, GFT-HQ reduces both the amplitude and the mean of this cycle. This leads to a more stable error profile that persists across all forecast leads.

This benefit is not confined to a particular time of day or season. GFT-HQ achieves the lowest Mean Absolute Error (MAE) in all diurnal periods (Morning, Afternoon, Evening, Night) and across all four seasons. Notably, its performance gains are largest during evening and nighttime hours, which are operationally critical periods of high energy demand and icing risk. While absolute forecast errors are naturally highest in winter and lowest in summer, GFT-HQ's performance advantage over ECMWF-IFS and GFT remains constant year-round.

Operationally, this multi-faceted consistency is highly valuable. The reduced diurnal and seasonal error volatility translates directly into more reliable and trustworthy day-ahead wind power forecasts, supporting more efficient grid management and resource planning throughout the year.

\subsubsection{Temperature and precipitation forecasting}
\label{app:temp-precip-forecasting}
\begin{figure}
\centering
\begin{subfigure}[b]{0.99\textwidth}
    \centering
    \includegraphics[width=0.9\linewidth]{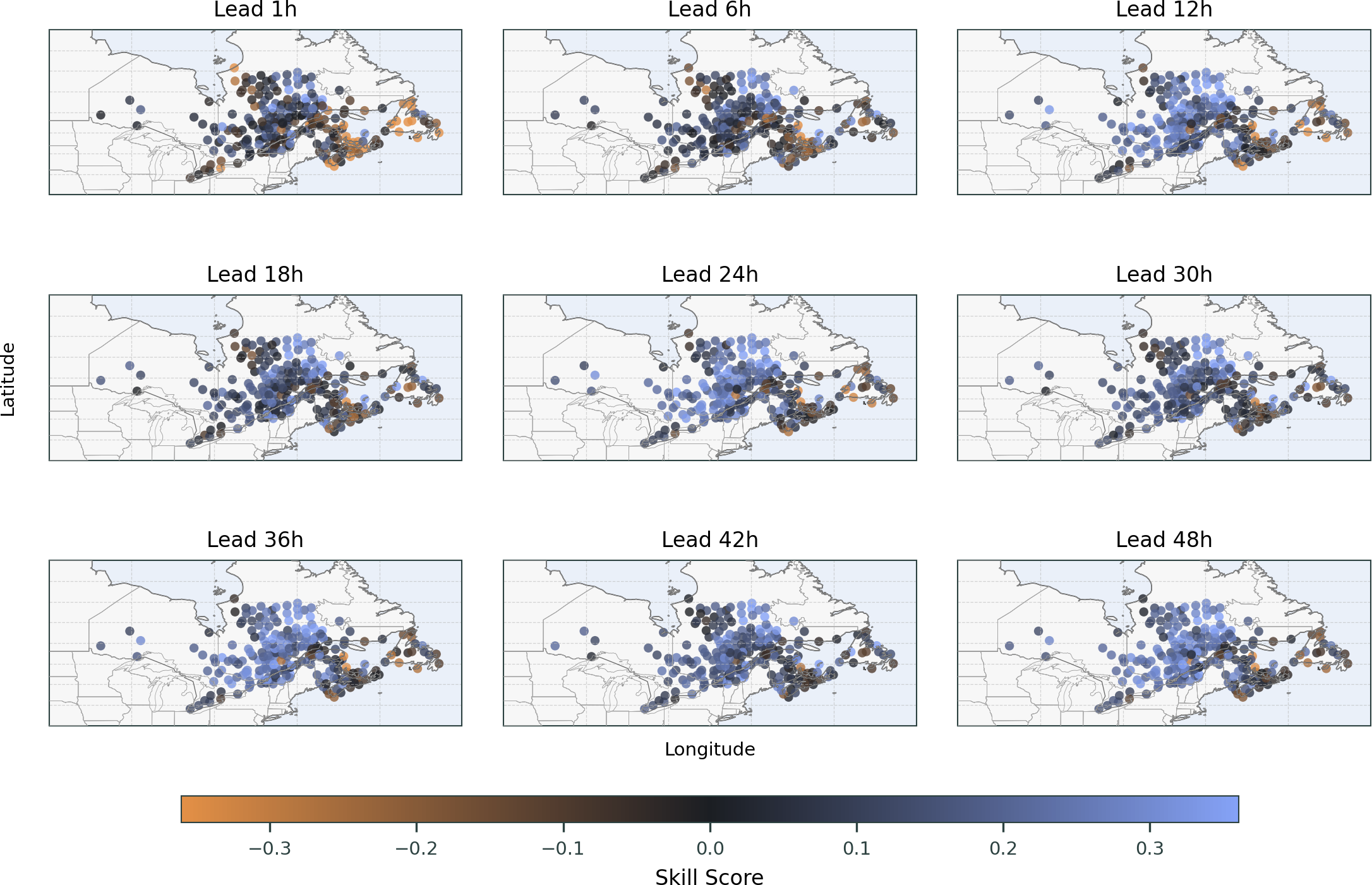}
    \caption{GFT-HQ temperature forecast skill improvement compared to ECMWF-IFS at different lead-times. }
    \label{fig:spatial_temperature_skill_gft_hq}    
\end{subfigure}
\begin{subfigure}[b]{0.99\textwidth}
    \centering
    \includegraphics[width=0.9\linewidth]{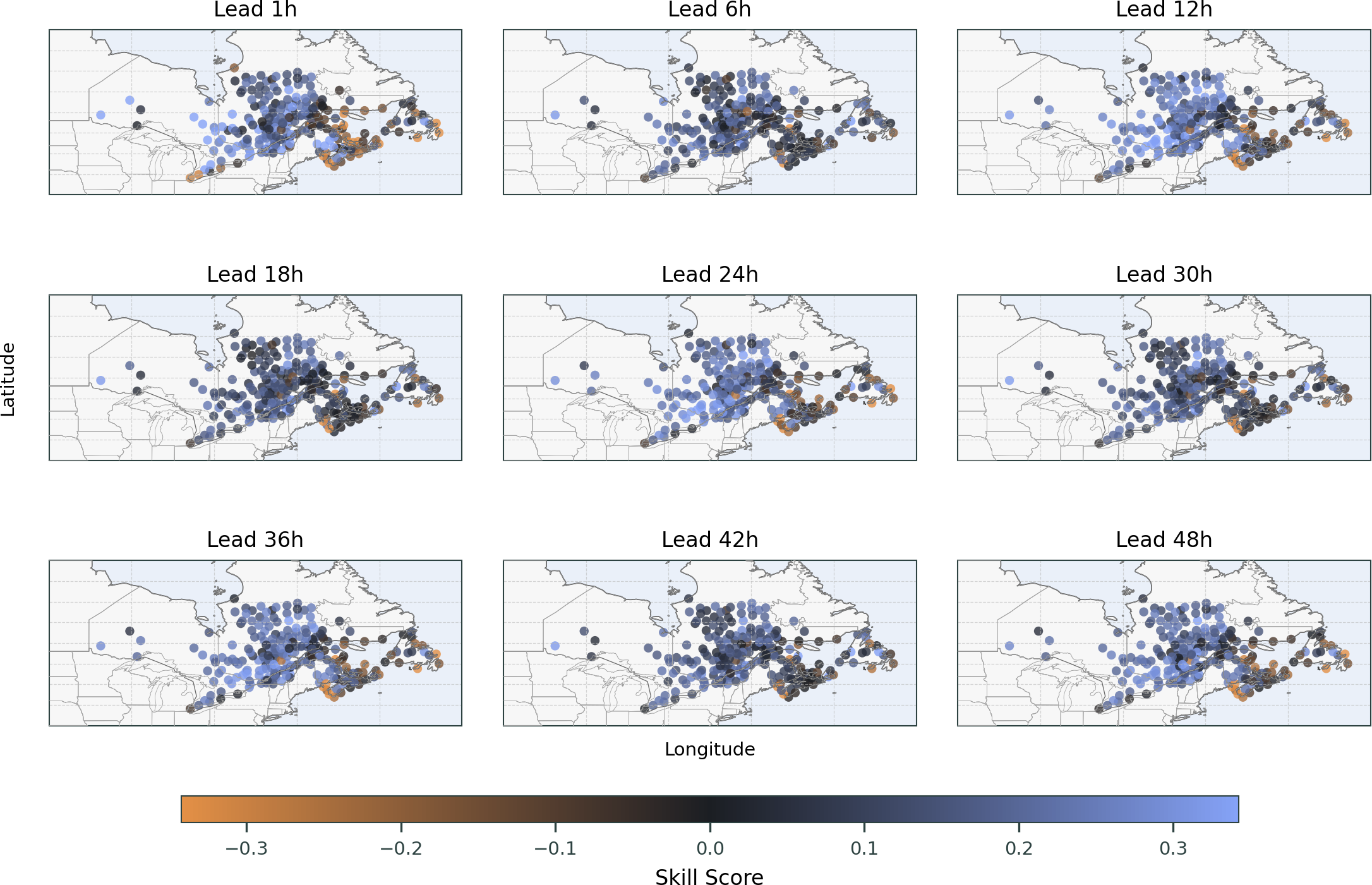}
    \caption{GFT temperature forecast skill improvement compared to ECMWF-IFS at different lead-times. }
    \label{fig:spatial_temperature_skill_gft}   
\end{subfigure}
\caption{Spatial distribution of GFT-HQ relative skill improvements compared to original GFT's skill against ECMWF-IFS. Higher score is better. GFT-HQ improved skill everywhere at longer lead-times, but it had a similar spatial distribution as the original GFT.}
\label{fig:temp_skill_maps}
\end{figure}

\paragraph{Spatial structure of temperature gains} 
\Cref{fig:temp_skill_maps} maps station-wise skill vs ECMWF-IFS (positive = lower error than IFS) at successive leads. Two patterns stand out:
\begin{itemize}
    \item Finetuning amplifies, it doesn’t relocate, the skill. The spatial pattern for GFT-HQ mirrors that of the untuned GFT, but with larger positive skill at virtually every lead. By 24-48 h the maps are dominated by positive (blue) stations for GFT-HQ, while the untuned GFT shows a similar but noticeably weaker footprint. This is expected given the finetuning setup here.
    \item Gains grow with lead time and remain geographically broad. Isolated neutral/negative pockets appear at short leads, but the fraction of improved stations increases monotonically with lead, consistent with \Cref{fig:snap_temperature_stations_improved}.
\end{itemize}

\textit{Operationally}, this means the temperature MAE reduction is system-wide rather than site-specific: load-forecast errors should fall across the control area, not just at a handful of stations, and the advantage persists into the day-ahead window where commitment and hedging decisions are made.

\begin{figure}
    \centering
\begin{subfigure}[b]{0.99\textwidth}
    \centering
    \includegraphics[width=0.9\linewidth]{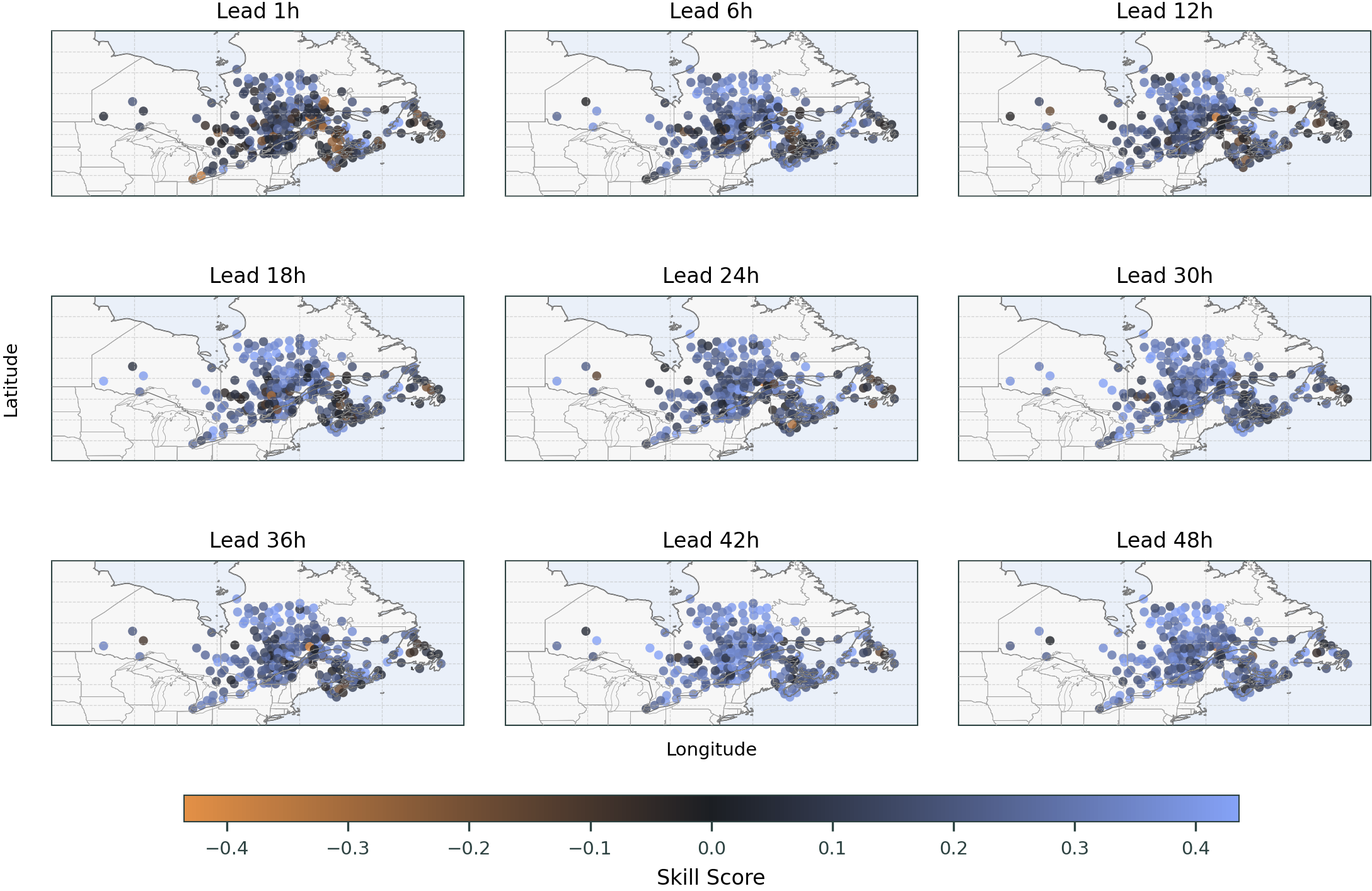}
    \caption{GFT-HQ hourly total precipitation forecast skill improvement compared to ECMWF-IFS at different lead-times. }
    \label{fig:spatial_precipitation_skill_gft_hq}    
\end{subfigure}
\begin{subfigure}[b]{0.99\textwidth}
    \centering
    \includegraphics[width=0.9\linewidth]{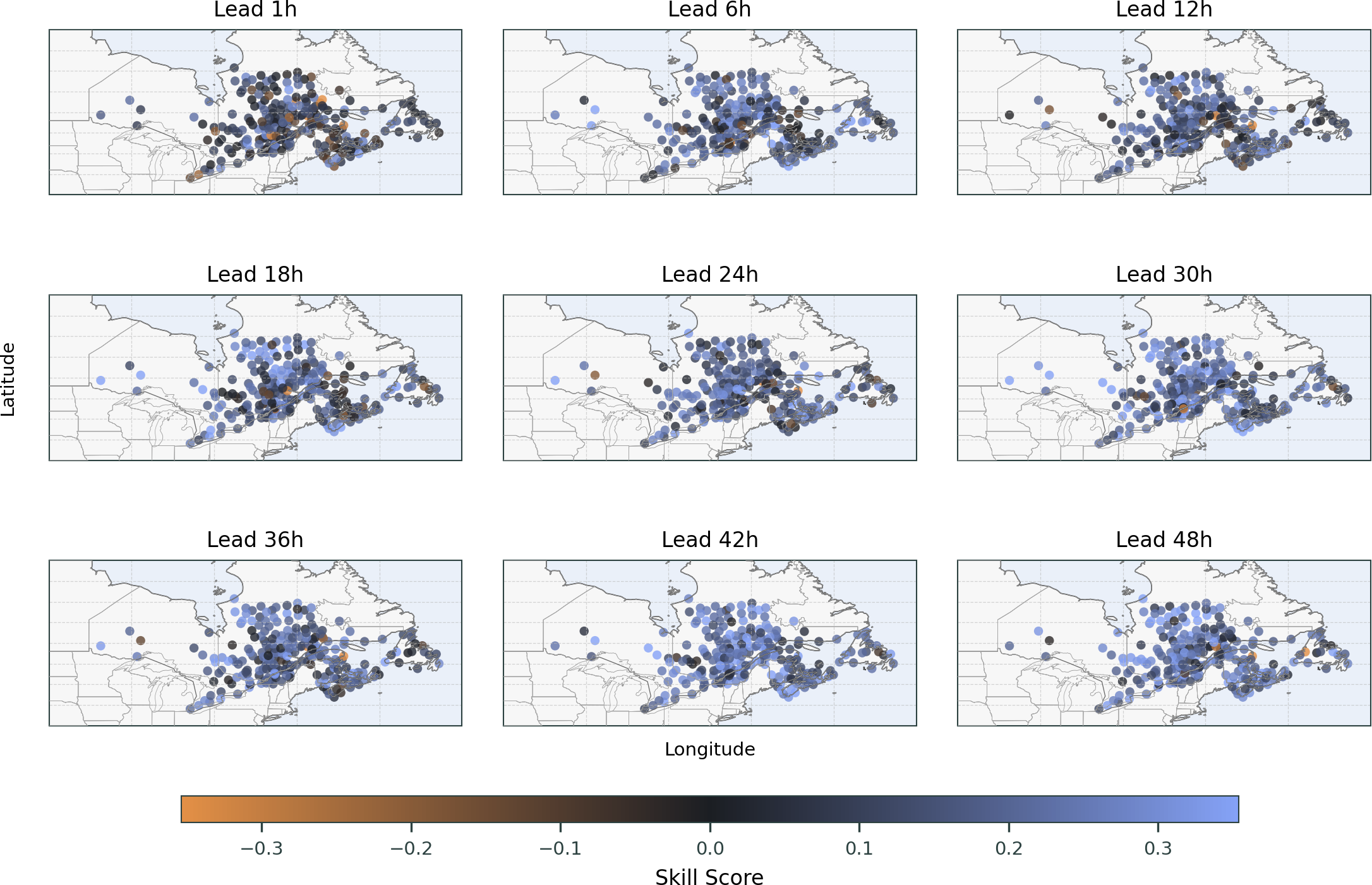}
    \caption{GFT hourly total precipitation forecast skill improvement compared to ECMWF-IFS at different lead-times. }
    \label{fig:spatial_precipitation_skill_gft}   
\end{subfigure}
    \caption{Spatial distribution of GFT-HQ's hourly precipitation skill improvements compared to original GFT's skill against ECMWF-IFS. Higher score is better. Finetuning increases the magnitude and coverage of positive skill at all leads.}
    \label{fig:tp_skill_maps}
\end{figure}

\paragraph{Spatial structure of precipitation gains}
\Cref{fig:tp_skill_maps} maps station-level skill vs. ECMWF-IFS for hourly total precipitation at successive leads (positive = lower error than IFS). The pattern is similar to improvements in temperature forecasts. Operationally, this means tighter basin-wide accumulation signals for hydro-inflow at the horizons HQ actually commits resources.

\begin{figure}[t]
\begin{subfigure}[b]{0.48\textwidth}
    \phantomsubcaption\label{fig:snap_temp_mae_morning} %
    \begin{overpic}[width=\textwidth]{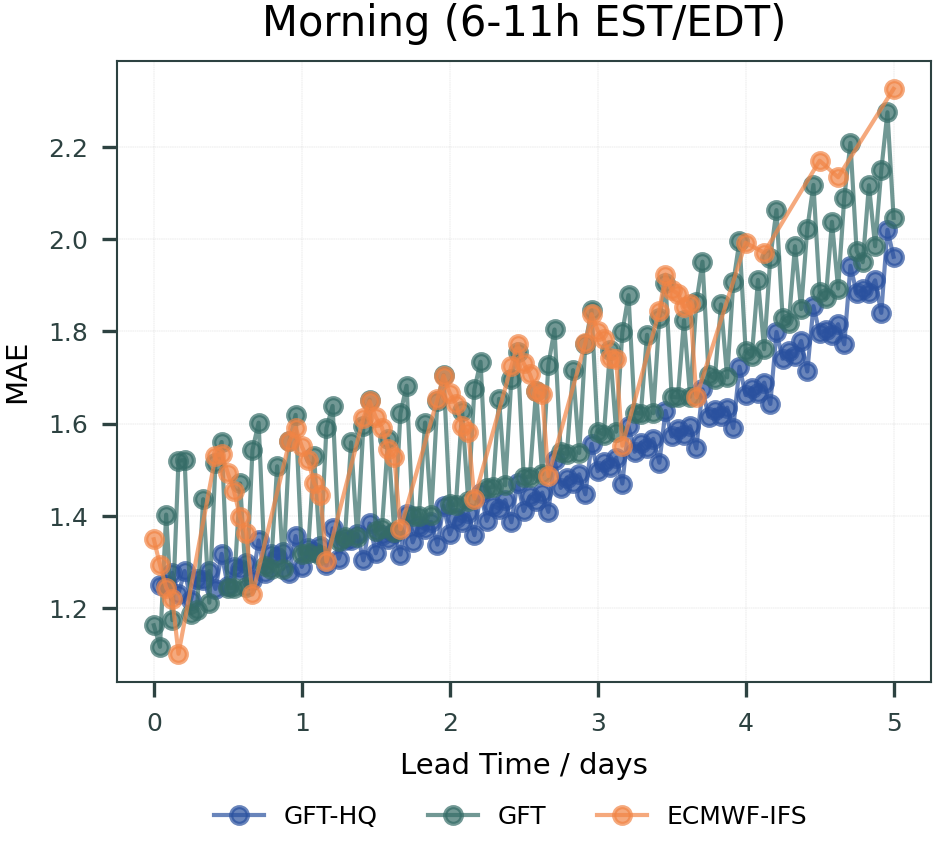}
      \put(2,82){\footnotesize\bfseries(\thesubfigure)}
    \end{overpic}
  \end{subfigure}
  \begin{subfigure}[b]{0.48\textwidth}
    \phantomsubcaption\label{fig:snap_temp_mae_afternoon}
    \begin{overpic}[width=\textwidth]{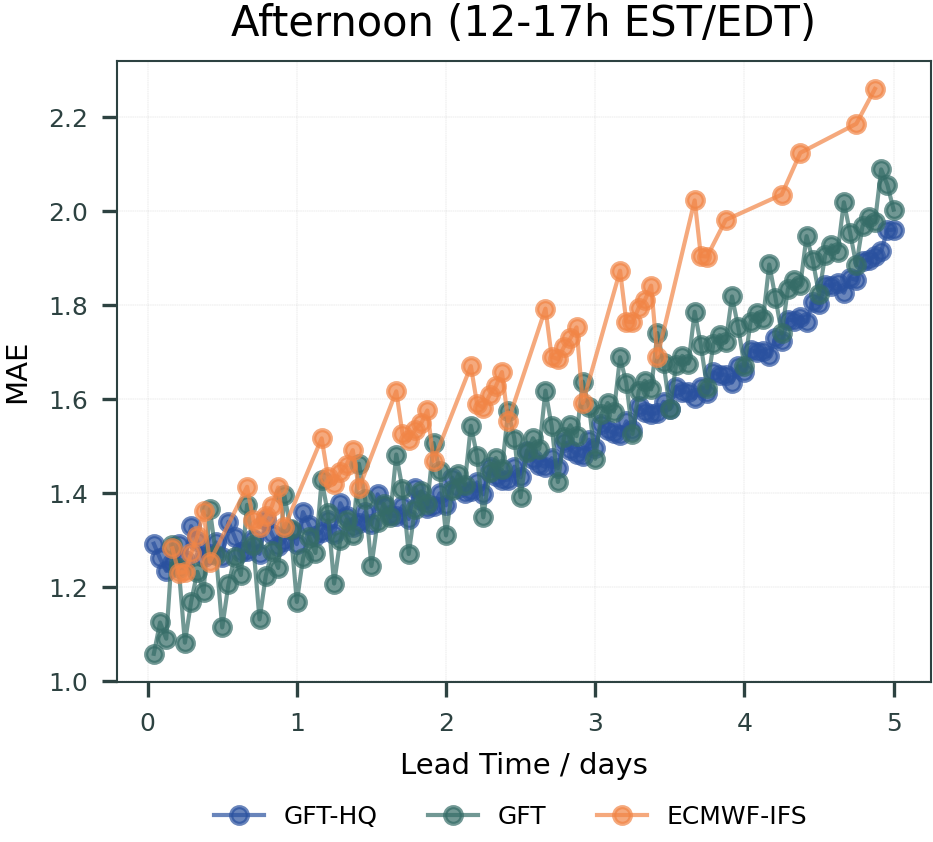}
      \put(2,82){\footnotesize\bfseries(\thesubfigure)}
    \end{overpic}
  \end{subfigure}
  \begin{subfigure}[b]{0.48\textwidth}
    \phantomsubcaption\label{fig:snap_temp_mae_evening} %
    \begin{overpic}[width=\textwidth]{figs/snap/temperature_Evening_mae.png}
      \put(2,82){\footnotesize\bfseries(\thesubfigure)}
    \end{overpic}
  \end{subfigure}
  \begin{subfigure}[b]{0.48\textwidth}
    \phantomsubcaption\label{fig:snap_temp_mae_night}
    \begin{overpic}[width=\textwidth]{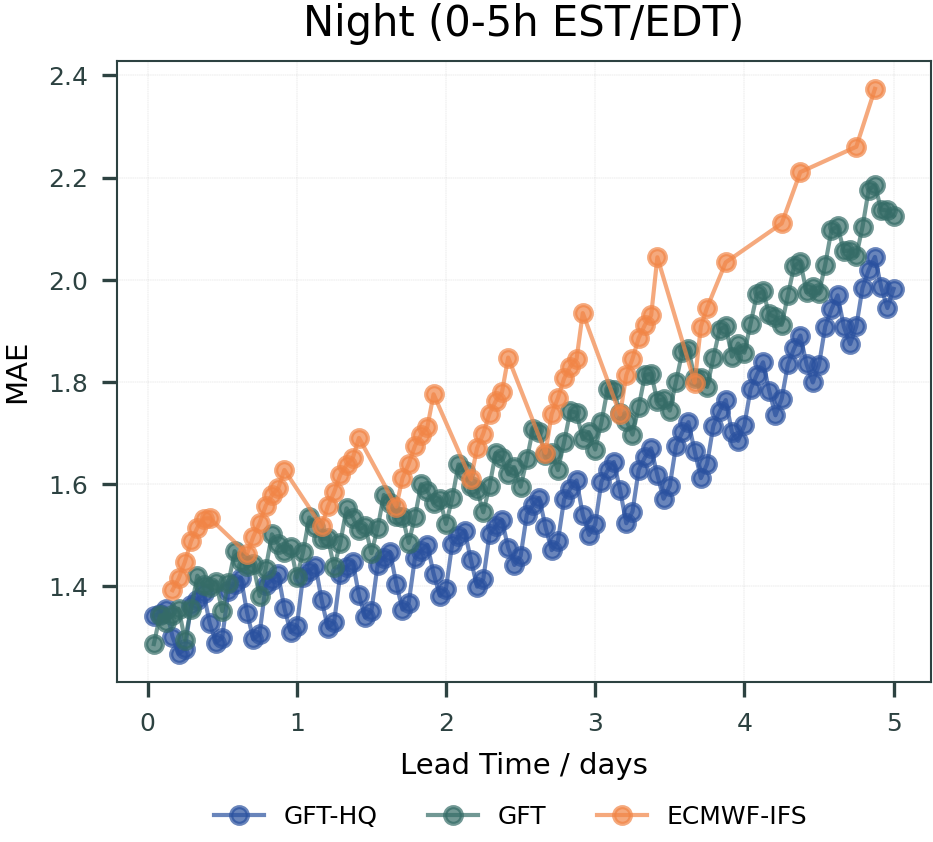}
      \put(2,82){\footnotesize\bfseries(\thesubfigure)}
    \end{overpic}
  \end{subfigure}  
  \caption{Surface temperature MAE by time of day. GFT-HQ outperforms other baselines under all conditions, except in the afternoons against raw GFT.}
\end{figure}

\begin{figure}[t]
\begin{subfigure}[b]{0.48\textwidth}
    \phantomsubcaption\label{fig:snap_temp_mae_spring} %
    \begin{overpic}[width=\textwidth]{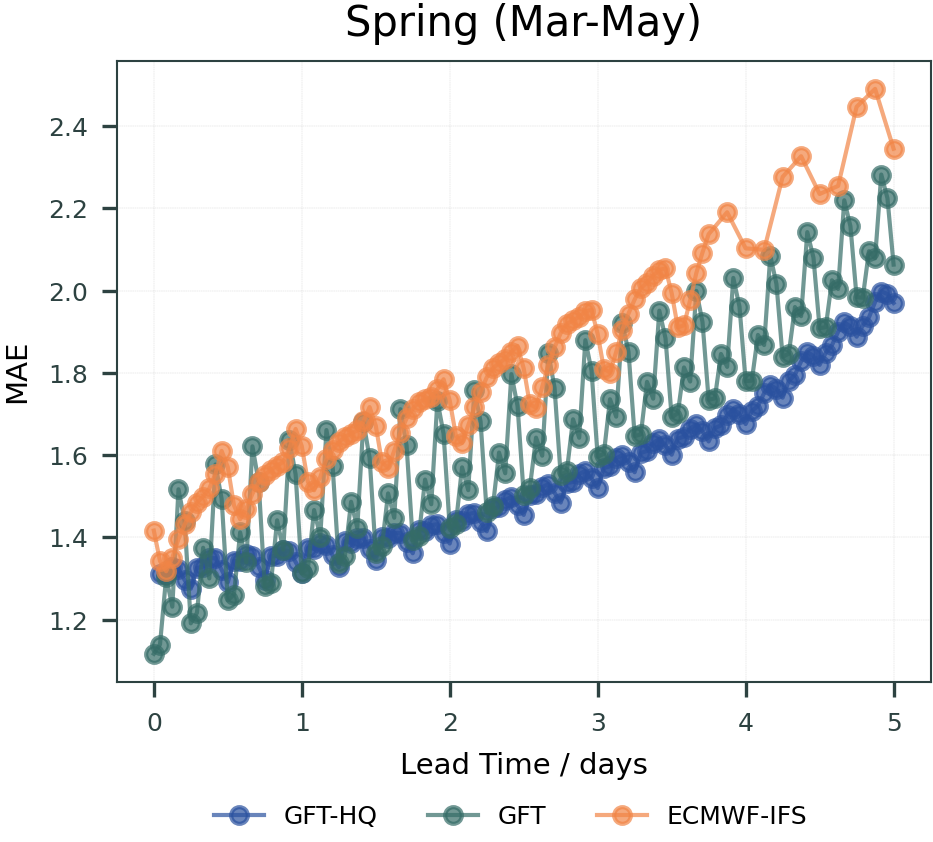}
      \put(2,82){\footnotesize\bfseries(\thesubfigure)}
    \end{overpic}
  \end{subfigure}
  \begin{subfigure}[b]{0.48\textwidth}
    \phantomsubcaption\label{fig:snap_temp_mae_summer}
    \begin{overpic}[width=\textwidth]{figs/snap/temperature_Summer_mae.png}
      \put(2,82){\footnotesize\bfseries(\thesubfigure)}
    \end{overpic}
  \end{subfigure}
  \begin{subfigure}[b]{0.48\textwidth}
    \phantomsubcaption\label{fig:snap_temp_mae_fall} %
    \begin{overpic}[width=\textwidth]{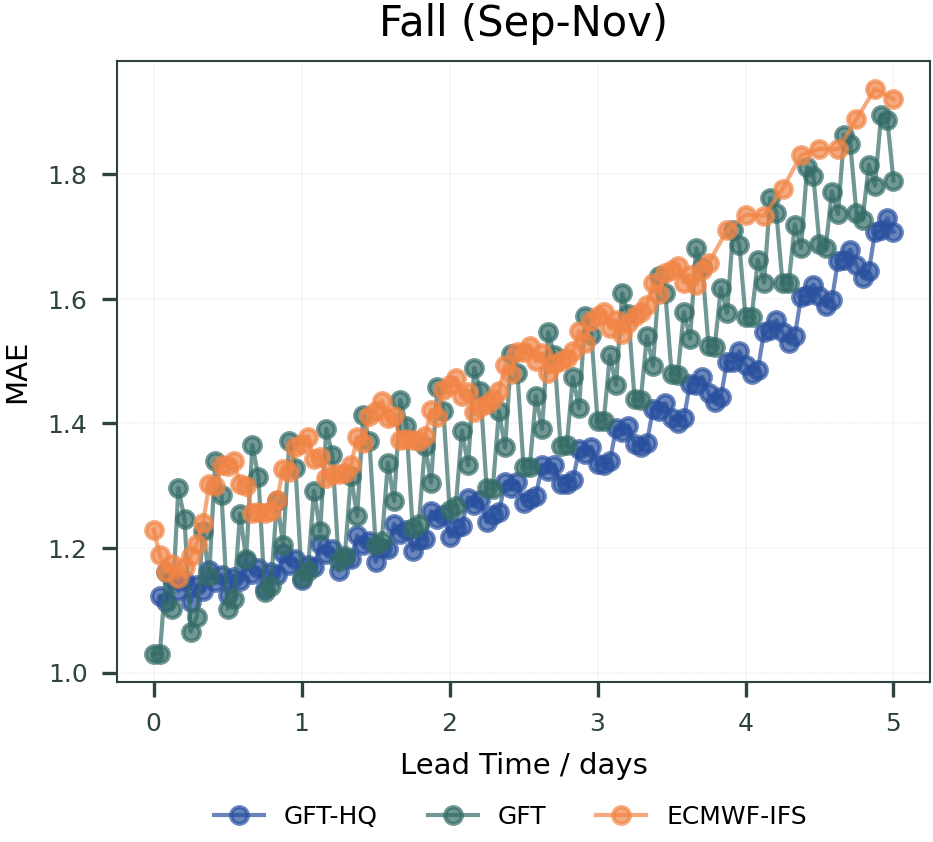}
      \put(2,82){\footnotesize\bfseries(\thesubfigure)}
    \end{overpic}
  \end{subfigure}
  \begin{subfigure}[b]{0.48\textwidth}
    \phantomsubcaption\label{fig:snap_temp_mae_winter}
    \begin{overpic}[width=\textwidth]{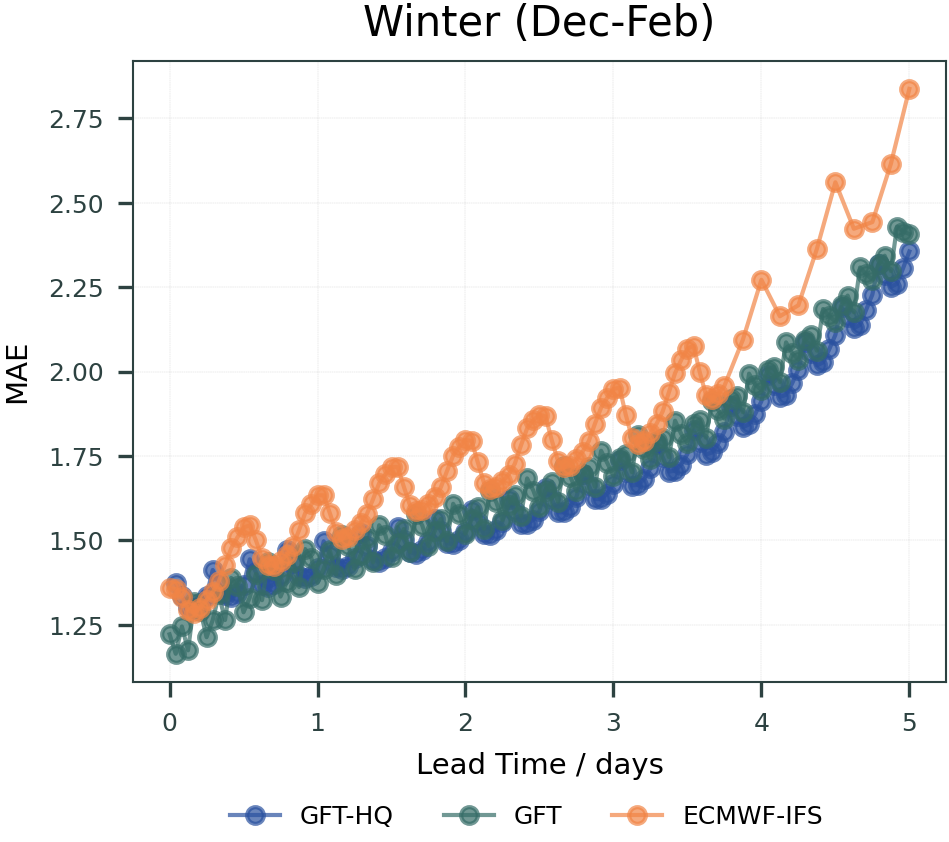}
      \put(2,82){\footnotesize\bfseries(\thesubfigure)}
    \end{overpic}
  \end{subfigure}  
  \caption{Surface temperature MAE by season. GFT-HQ outperforms other baselines under all conditions for 6-120h lead-times. However, the error patterns suggest that it could be helpful to oversample winter season during future fine-tuning.}
\end{figure}

\end{document}